%% file: main.tex
\documentclass[manuscript,screen,nonacm]{acmart} 

\AtBeginDocument{%
  \providecommand\BibTeX{{%
    \normalfont B\kern-0.5em{\scshape i\kern-0.25em b}\kern-0.8em\TeX}}}

\setcopyright{acmcopyright}
\copyrightyear{2023}
\acmYear{2023}
\acmDOI{XXXXXXX.XXXXXXX}

\acmJournal{CSUR}
\acmVolume{X}
\acmNumber{X}
\acmArticle{X}
\acmMonth{0}

\acmPrice{15.00}
\acmISBN{978-1-4503-XXXX-X/18/06}

\usepackage{amsmath} 
\usepackage{float} 
\usepackage{array} 
\usepackage{enumitem} 
\usepackage{hyperref} 
\usepackage{comment}
\usepackage{ragged2e} 
\usepackage{multirow} 

\usepackage[absolute,overlay]{textpos}
\usepackage{tikz}
\textblockorigin{0cm}{0cm}

\begin{document}

\begin{textblock}{15}(1.75,24)
    \begin{tikzpicture}
        \draw[line width=1pt, color=black] (0,0) rectangle (16,1.5); 
        \node at (8,0.75) { 
            \begin{minipage}{15.5cm} 
                © {Pedro Miguel Moás and Carla Teixeira Lopes} {2023}. This is the author's version of the work. It is posted here for your personal use. Not for redistribution. The definitive Version of Record was published in ACM Computing Surveys, \url{https://dx.doi.org/10.1145/3625286}
            \end{minipage}
        };
    \end{tikzpicture}
\end{textblock}

\title[Automatic Quality Assessment of Wikipedia Articles - A Systematic Literature Review]{Automatic Quality Assessment of Wikipedia Articles - A Systematic Literature Review}


\author{Pedro Miguel Moás}
\email{pmmoas@fe.up.pt}
\affiliation{
  \institution{Faculdade de Engenharia da Universidade do Porto}
  \streetaddress{R. Dr. Roberto Frias, s/n}
  \city{Porto}
  \country{Portugal}
  \postcode{4200-465}
}

\author{Carla Teixeira Lopes}
\email{ctl@fe.up.pt}
\affiliation{
  \institution{Faculdade de Engenharia da Universidade do Porto, INESC TEC}
  \streetaddress{R. Dr. Roberto Frias, s/n}
  \city{Porto}
  \country{Portugal}
  \postcode{4200-465}
}


\include{vars}

\include{abstract}

\include{ccs}

\keywords{Wikipedia, Quality Assessment, Information Quality} 


\maketitle

\input{sections/section1-intro}
\input{sections/section2-background}
\input{sections/section3-methodology}
\input{sections/section4-res-overview}
\input{sections/section5-res-ml}
\input{sections/section6-res-features}

\input{sections/section7-discussion}

\input{sections/section8-conclusion}

\bibliographystyle{ACM-Reference-Format}
\bibliography{main}

\appendix

\end{document}

%% file: vars.tex
\def\RQONE{What are the most commonly used methods for the automatic quality assessment of Wikipedia articles?}
\def\RQTWO{How can machine learning be best applied to predict article quality, and how do different approaches compare?}
\def\RQTHREE{What are the most common article features and quality metrics used to evaluate article quality in Wikipedia? How do these features compare, and how do they affect the performance of automatic assessment methods?}
\def\RQFOUR{Which common themes and gaps are there in the literature concerning this topic, and how can existing studies be improved to increase the adoption of automatic methods for the quality assessment of Wikipedia?}

\def\PUBTOTAL{149}


\def\PUBTYPICALMETHODS{102}
\def\PUBFEATSCORRELATES{20}
\def\PUBNONENGLISH{35}

\def\PUBONELANG{113}
\def\PUBONELANGML{66}
\def\PUBONELANGNOML{47}

\def\PUBMETRICS{71} 
\def\PUBMETRICBASED{31} 

\def\ALLFEATURES{1786} 
\def\FEATURES{321} 
\def\PUBFEATURES{104} 

\def\PUBML{81}
\def\PUBNOML{68}
\def\PUBCL{65}
\def\PUBDL{20}
\def\PUBCLDL{4}
\def\MLEXPERIMENTS{215}
\def\PUBBADMETRICS{13}

\def\MEANDATASIZEML{$81,203$}
\def\MEANDATASIZENOML{$1,058,111$}

%% file: abstract.tex
\begin{abstract}

Wikipedia is the world's largest online encyclopedia, but maintaining article quality through collaboration is challenging. Wikipedia designed a quality scale, but with such a manual assessment process, many articles remain unassessed.
We review existing methods for automatically measuring the quality of Wikipedia articles, identifying and comparing machine learning algorithms, article features, quality metrics, and used datasets, examining \PUBTOTAL\ distinct studies, and exploring commonalities and gaps in them. The literature is extensive, and the approaches follow past technological trends. However, machine learning is still not widely used by Wikipedia, and we hope that our analysis helps future researchers change that reality.

\end{abstract}

%% file: ccs.tex
\ccsdesc[500]{Computing methodologies~Natural language processing}
\ccsdesc[500]{Information systems~Wikis}
\ccsdesc[300]{Computing methodologies~Machine learning}
\ccsdesc[300]{Applied computing~Document management and text processing}

%% file: sections/section1-intro.tex
\section{Introduction} \label{sec:intro}

Wikipedia is the largest and most well-known online encyclopedia and has kept its growing pace for years. As of April 2023, it contains over 6.6 million English articles~\cite{Web_WikipediaSize}, with versions across a list of 321 active languages~\cite{Web_WikipediaList}.

Not only is Wikipedia free, but it is also fully managed by human volunteers, averaging contributions at a rate of 5.7 edits per second during 2022~\cite{Web_WikistatsWikimedia}. Its reader base is growing steadily, considering that, last year, Wikipedia totaled close to 280 billion page views across 2 billion unique devices~\cite{Web_WikistatsWikimedia}. 
Some studies even show that most search engines frequently include Wikipedia pages in their results. According to Vincent and Hecht~\cite{Vincent2021_WikiGoogle}, 80\% of \emph{common} (frequent) queries and 70\% of \emph{trending} queries (news/events) return results from Wikipedia on the first page, when tested with search engines like Google, Bing, and DuckDuckGo.


The fully collaborative aspect of Wikipedia brings its own set of challenges too, as the lack of centralized authority over the editors makes it challenging to ensure quality throughout the website. Only 8.1\% of edits are reverted~\cite{Web_WikistatsWikimediaReverts}, showing that vandalism and the so-called \emph{revert wars} are relatively uncommon, but it is still crucial to ensure the improvement of low-quality articles. 

Another issue is the substantial quality discrepancy between English Wikipedia and its other versions. First, each non-English version covers a much smaller amount of articles~\cite{Web_WikipediaList}. Also, they are often much more incomplete as well. Roy et al.~\cite{Roy2020_WikiAssymetry, Roy2022_WikiAssymetry} demonstrate that English articles from Wikipedia are usually longer than their translations. They determined that German articles are, on average, 30\% shorter than English ones, and for Spanish articles, that value increases to 47\%, but there are still some English articles that are much less complete than their non-English counterparts. Couto and Lopes~\cite{Couto2021_lr161} have also shown this quality discrepancy, although only focused on health-related articles. They used a set of metrics to determine that English articles show the best values for quality, ranking much higher than other idioms. 

To better monitor and help maintain the quality of the website, Wikipedia designed a quality scale that aims to rate articles within one of 9 possible grades, which go from the most incomplete documents (Starts and Stubs) to the most comprehensive, well-written articles (Featured Articles)~\cite{Web_WikipediaContentAss}. However, the majority of Wikipedia is made up of lower-quality articles, with Starts and Stubs accounting for more than 80\% of its English content. In comparison, the share of Featured Articles and Good Articles is around 0.7\%~\cite{Web_WikipediaContentAss}. Nonetheless, these values are not meant to be taken as official ratings but instead for internal use by the contributors. Besides, not every English article is rated, and the non-English versions of Wikipedia that also assess their content will have different quality scales, evaluated with other criteria. For those reasons, Wikipedia users lack a consistent and transparent method for determining the quality of articles.


Our goal is to review proposed methods for automatically measuring the quality of Wikipedia articles. There already exist a couple of studies sharing a similar objective~\cite{Fahimnia2022_lr118, Bassani2019_lr359}, but they mostly focus on used article features and do not provide a detailed and exhaustive overview of the existing publications. We believe it would be beneficial to dive deeper into the subject of automatic assessment of Wikipedia quality, so we conduct a systematic literature review to assess the state of the art within that topic, examining and comparing existing approaches used to automatically measure article quality. Specifically, we analyzed machine learning methods, article features, quality metrics, datasets, and other common aspects of these approaches, such as multilingual assessment and data visualization/explanation tools for supporting the reader and editor community. With this review, we aim to provide a starting point for future work that aims to understand Wikipedia quality and design automatic methods for measuring it. 


We divided this article into eight sections. After this introduction in Section~\ref{sec:intro}, Section~\ref{sec:background} provides some insight about Information Quality. Section~\ref{sec:method} details our methodology for the systematic review, and we list its results in the following sections: Section~\ref{sec:res-overview} overviews the included papers and the methods they use, Section~\ref{sec:res-ml} summarizes used machine learning approaches, and Section~\ref{sec:res-features} analyses applied article features and quality metrics. We summarize and discuss our findings in Section~\ref{sec:discussion}, answering the defined research questions. Finally, we conclude this paper in Section~\ref{sec:conclusions}, where we reflect on our study and examine future work possibilities.

%% file: sections/section2-background.tex
\section{Information Quality} \label{sec:background}

It is important to design our definition of quality. Hence, we must first answer: \emph{What is quality? Can it be objectively quantified?} Information Quality (IQ) is an extraordinarily researched topic, and there exist numerous attempts to provide a way to calculate it. Lee et al.~\cite{Lee2002_AIMQ} break down the measurement of Information Quality into 15 properties, including \emph{Accessibility}, \emph{Believability}, \emph{Interpretability}, \emph{Objectivity}, \emph{Reputation}, and \emph{Timeliness}. Some of these aspects are much easier to assess than others. However, with recent developments in Natural Language Processing (NLP), some works already attempt to evaluate more complex topics such as bias~\cite{Hube2018_BiasAssessment}, neutrality~\cite{AlKhatib2012_POVAssessment}, trustworthiness~\cite{Kittur2008_Trust1, Kuznetsov2022_Trust2, Adler2008_Trust3, Zend2006_Trust4, Dondio2006_Trust5}. Although these studies are an inspiration for authors attempting to tackle the topic of this review, we do not intend to include them unless they specifically propose methods for automatically predicting the quality of articles.

Wikipedia has its definition of quality, too. For instance, the English Wikipedia content assessment guidelines~\cite{Web_WikipediaContentAss} indicate that the most outstanding articles must be well-written, comprehensive, well-researched, and follow their style guidelines~\cite{Web_WikipediaStyle}, which relate to the IQ properties defined by Lee et al.~\cite{Lee2002_AIMQ}. However, that definition may vary even within Wikipedia: according to Jemielniak and Wilamowski~\cite{Jemielniak2017_lr262}, not all language cultures share the same understanding of quality, which is a vital aspect to consider when designing a multilingual solution for quality assessment.

Overall, quality is a subjective property, so it is difficult to design an objective definition for it. However, there are certainly measurable characteristics that people often relate to outstanding quality, and we plan to determine them and their correlation with excellence in written documents, better understanding which are the most effective methods for predicting it.

%% file: sections/section3-methodology.tex
\section{Methodology} \label{sec:method}

This systematic review aims to answer the following research questions:

\begin{enumerate}[itemsep=0pt]
  \item[RQ1.] \RQONE\
  \item[RQ2.] \RQTWO\
  \item[RQ3.] \RQTHREE\
  \item[RQ4.] \RQFOUR\
\end{enumerate}

\begin{figure}[ht]
    \centering
    \includegraphics[width=0.9\textwidth]{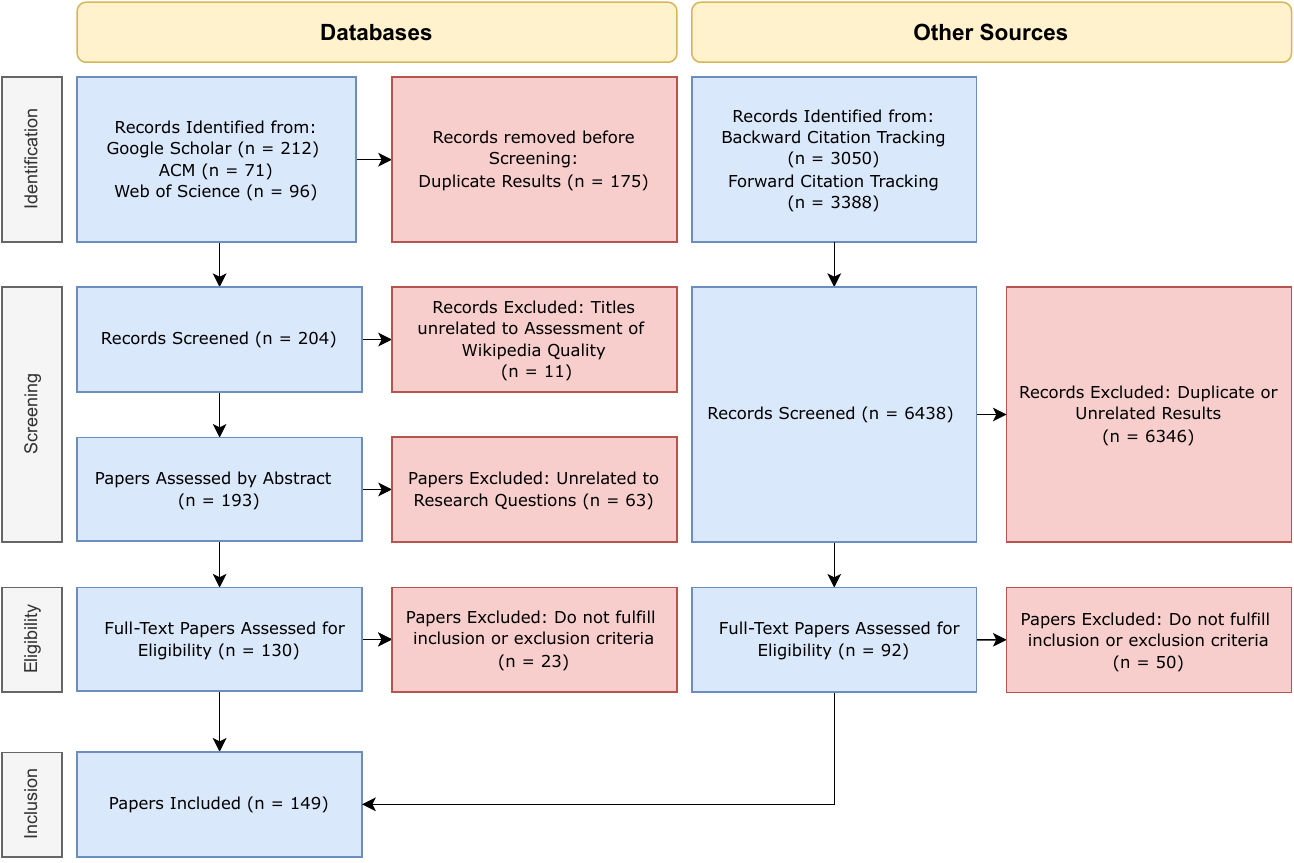}
    \caption{An overview of the selection process of our review}
    \label{fig:prisma}
\end{figure}

We guided our selection process by the PRISMA statement \cite{Page2021_PRISMA1, Page2021_PRISMA2}, which defines a set of guidelines for conducting systematic literature reviews. Our selection included two main selection stages: \textbf{Database Querying} and \textbf{Citation Tracking}. Figure~\ref{fig:prisma} outlines the selection process by detailing the number of publications included in each phase. Initially, we selected a set of records using research databases, and next, we conducted citation tracking on a subset of the initial selection.

\subsection{Selection through Database Querying}

Our initial selection stage comprises 4 phases: Identification, Screening, Eligibility, and Inclusion. 

\subsubsection{Identification}
We considered three primary data sources: Google Scholar, ACM Digital Library, and Web of Science. In all of them, we retrieved all results containing "Wikipedia" and "quality" in its title. Searching in the title significantly reduces the number of results (for instance, reduces Google Scholar results by ~99.97\%), allowing the screening of all the retrieved results. We present the exact query and number of results for each database in Table~\ref{tab:queries}.

\begin{table}[ht]
    \caption{Search queries submitted in each database}
    \label{tab:queries}
    \centering
    \begin{tabular}{
        >{\centering\arraybackslash}p{0.2\textwidth}
        p{0.425\textwidth}
        >{\centering\arraybackslash}p{0.1\textwidth}
    }
        \hline
            \multicolumn{1}{c}{\textbf{Database}} &
            \multicolumn{1}{c}{\textbf{Query}} &
            \multicolumn{1}{c}{\textbf{Results}} \\ 
        \hline
            Google Scholar$^{(1)}$ & 
            allintitle:``wikipedia'' ``quality'' &  
            212 \\  
        \hline
            ACM & 
            [Title: wikipedia] AND [Title: quality] AND \newline [E-Publication Date: (01/01/2001 TO 01/31/2023)] &
            71 \\
        \hline
            Web of Science$^{(1)}$ & 
            Wikipedia quality$^{(2)}$ & 
            96 \\
        \hline
    \end{tabular}
    \\
    $^{(1)}$We applied a date range of 2001-2023 \\
    $^{(2)}$We applied this query on the title field \\
\end{table}

All database queries were run on January 24\textsuperscript{th}, 2023, and were restricted to a date range of 2001-2023. We picked a lower limit of 2001 because Wikipedia was launched in that year~\cite{Web_Wikipedia}, so we would unlikely find related articles from an earlier date.

Our Identification phase returned 379 results, as shown in Table~\ref{tab:queries}. We discarded 175 records as duplicates.

\subsubsection{Screening}

Next, publication titles were assessed for possible relevance within the research area. All results that appeared at least marginally related to the quality assessment of Wikipedia proceeded to the next phase. Nearly all results moved forward, as our query was already somewhat strict. From the 204 non-duplicate titles, 193 were considered possibly relevant for our research. For instance, Salutari et al.'s study~\cite{Salutari2019_ExcludedExample} was one of the results our query retrieved, but we deemed its title (``A Large-Scale Study of Wikipedia Users’ Quality of Experience'') unrelated to our research topic.

In this phase, we also excluded results that did not respect the usual research article format. This includes theses, dissertations, and technical reports, among others. We opted to include pre-prints to avoid the exclusion of potentially insightful papers. 

Finally, we scanned the papers' abstracts, focusing on the research questions. Only studies that propose automatic methods for measuring Wikipedia quality were included. This phase excluded publications experimenting with manual quality assessment approaches within specific sub-fields (e.g., Health) and studies whose abstracts we could not find. We advanced 130 studies to the next phase.

\subsubsection{Eligibility}

In the Eligibility phase, we first defined clear inclusion and exclusion criteria and assessed all the manuscripts that reached it. These criteria are described in Tables~\ref{tab:inclusion} and~\ref{tab:exclusion}, and resulted in the exclusion of 23 results. We manually excluded all the publications that did not meet any inclusion criteria and publications that met at least one exclusion criterion. 

\begin{table}[ht]
    \caption{Inclusion Criteria}
    \label{tab:inclusion}
    \centering
    \begin{tabular}{p{0.05\textwidth} p{0.8\textwidth}}
        \hline
        \textbf{ID} & \textbf{Criteria} \\ 
        \hline
        I1 & Paper discusses machine learning approaches to predict information quality. \\  
        \hline
        I2 & Paper discusses possible features or metrics to assess information quality. \\
        \hline
    \end{tabular}
\end{table}

\begin{table}[ht]
    \caption{Exclusion Criteria}
    \label{tab:exclusion}
    \centering
    \begin{tabular}{p{0.05\textwidth} p{0.8\textwidth}}
        \hline
        \textbf{ID} & \textbf{Criteria} \\ 
        \hline
        E1 & Paper does not discuss the assessment or prediction of the quality of a collaborative network. This includes studies that only discuss approaches to assess vandalism, controversy, or trust. \\
        \hline
        E2 & Paper discusses manual approaches to assess article quality, as opposed to automatic ones. \\
        \hline
        E3 & Paper is not in the English language. \\
        \hline
    \end{tabular}
\end{table}

\subsubsection{Inclusion}

In this phase, we run a serious analysis of each paper to collect all information relevant to our study, as we will describe in Section~\ref{subsec:data_collection}. Overall, this initial selection stage included 107 papers.

\subsection{Selection through Citation Tracking}

To minimize the probability of excluding relevant articles, we run citation tracking~\cite{Haddaway2021_CitationChaser}, searching through the references (backward tracking) and citations (forward tracking) of all included articles to identify potentially useful results. 

Naturally, this procedure directly scales with the number of included articles and the respective number of references and citations. We determined that tracking the entire result set would be impractical, so we decided to only perform backward and forward tracking on the most relevant papers. We assessed relevance using a systematized scoring process, where we assign an integer value from 0 to 10 based on four questions, as listed in Table~\ref{tab:relevance_q}. We performed citation tracking on all results yielding a global relevance score of 4 or higher. 

\begin{table}[ht]
    \caption{Citation Tracking: Relevance Scoring Questions}
    \label{tab:relevance_q}
    \centering
    \begin{tabular}{c m{0.55\textwidth} m{0.35\textwidth}}
        \hline
        \textbf{ID} & \textbf{Question} & \textbf{Possible Scores$^*$} \\ 
        \hline
        Q1 & Does the study focus on the topic of automatic quality assessment of Wikipedia? & From 0 (strongly disagree) to 3 (strongly agree) \\
        \hline
        Q2 & Does the study describe and compare multiple ML approaches? & 0 ($MLExp = 0$), 1 ($1\leq MLExp < 4$), 2 ($4\leq MLExp < 7$), and 3 ($MLExp\geq7$)\\
        \hline
        Q3 & Does the study describe and compare multiple article features and quality metrics? & 0 ($FT = 0$), 1 ($1\leq FT < 15$), 2 ($15\leq FT < 50$), and 3 ($FT\geq50$) \\
        \hline
        Q4 & Does the study focus on an article language that's not English? & 0 (No), 1 (Yes) \\
        \hline
        
    \end{tabular}
    \\ \vspace{0.1cm}
    \footnotesize
    \justifying
    $^*$ MLExp corresponds to the number of used machine learning experiments and FT to the number of used features.
\end{table}

For each article, we manually checked the titles and abstracts of each reference and citation (we obtained citation data from Google Scholar in March of 2023), applying the same criteria used during the first Screening phase. 

All relevant results transitioned to the Eligibility phase directly, therefore, will be assessed for inclusion and may end up being re-tracked, given a high enough relevance score. Overall, we performed this process on 92 different publications, which led to the inclusion of 42 new publications. Our systematic literature review included a total of \PUBTOTAL\ studies. 

\subsection{Data Collection} \label{subsec:data_collection}

Throughout every phase of the selection process, we systematically logged all the data we collected and produced.

Initially, we store the title of every record we gathered during the first Identification phase and assign them a numeric identifier. We also stored abstracts of publications that advanced to that sub-step of the Screening phase. Due to the substantial amount of inspected references and citations (6438), we did not keep any metadata for publications excluded during the Screening phase of Citation Tracking. 

We extracted most of the information during the Inclusion phase. We began by collecting relevant metadata of the \PUBTOTAL\ studies, such as the title, abstracts, keywords, authors, and year of publication. We then gathered study data, namely machine learning algorithms and respective performance, used article features and quality metrics, and dataset information.

All the information we collected is available in a research data repository~\cite{Moas2023_Dataset}, allowing readers to consult all the raw information we aggregated to display the results. We also provide a spreadsheet version of the dataset, similar to how we present it in this article, simplifying access for those who prefer not to handle the raw data directly.

%% file: sections/section4-res-overview.tex
\section{Overview of Included Articles} \label{sec:res-overview}

This section provides an overview of the \PUBTOTAL\ included papers~\input{sections/results/allcits} analyzing used methods and assessing metadata attributes, like publication venues, citation count, authors, and keywords.

\subsection{Methods}

Most papers (\PUBTYPICALMETHODS\ out of \PUBTOTAL) follow one of these quality assessment strategies: classical learning (CL) models trained with article features, deep learning (DL) methods using full text or features, and metric-based approaches (MB). Many publications also study the correlation of specific features with quality: although they are not concrete automatic methods for quality prediction, we still consider them relevant for the purpose of this study. We summarize this information in Table~\ref{tab:methods}.

\input{sections/results/overview/methods}

\subsubsection{Actionable Models and Visualization Tools}

Designing an effective quality model for Wikipedia greatly assists Wikipedia users, by allowing easier identification of the best and worst articles. However, this does little for editors who wish to improve them. In the context of Explainable AI~\cite{Xu2019_ExplainableAI} it is important to create solutions that also suggest improvement paths, like the actionable model proposed by Warncke-Wang~\cite{Warncke-Wang2013_lr13}. Some studies propose visualization tools that help solve this aspect. For instance, WikiRank~\cite{Wecel2015_lr34} provides quality information and popularity stats of articles across many languages. Other studies~\cite{Chevalier2010_lr65, Sciascio2017_lr83, Dalip2011_lr111} share a similar goal, although their solutions are much less thorough.
 
\subsubsection{Multi-language Assessment}

\begin{figure}[htbp]
    \centering
    \includegraphics[width=0.8\textwidth]{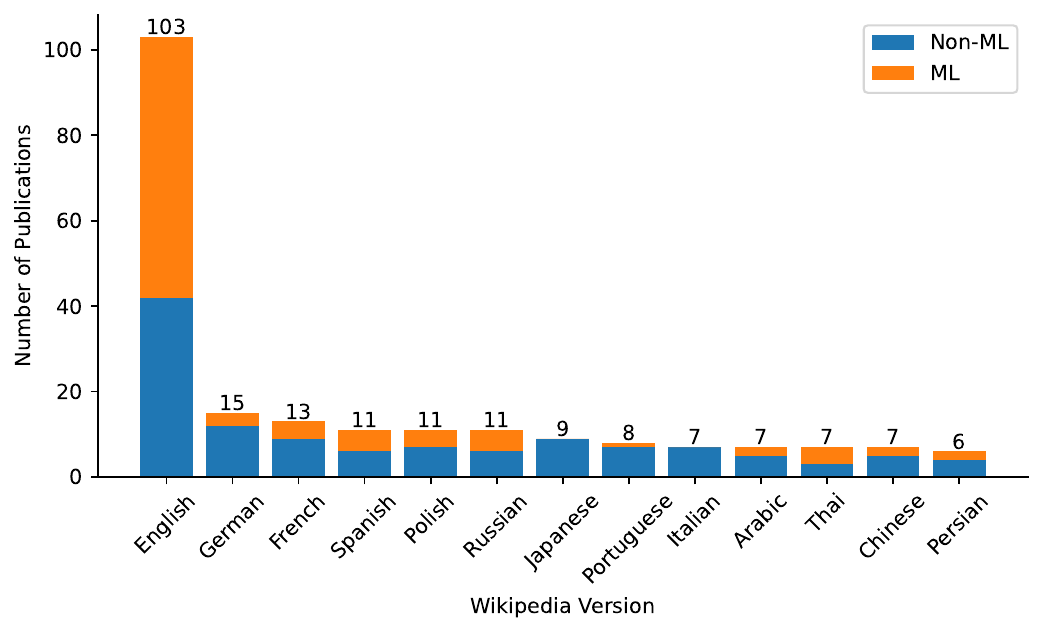}
    \caption{Most commonly studied Wikipedia versions in included publications (only versions with more than 5 publications are shown)}
    \label{fig:languages}
\end{figure}

As explained in Section~\ref{sec:intro}, article quality varies significantly across different Wikipedia versions, so we tried to understand to what extent authors have studied quality assessment in multiple languages. Figure~\ref{fig:languages} shows that authors mostly focus on the English Wikipedia, but there are still some publications that consider other languages, occasionally within a machine learning context. We also discovered that \PUBNONENGLISH\ papers exclusively consider non-English Wikipedias~\cite{Wohner2009_lr10, Druck2008_lr15, Cusinato2009_lr33, Ingawale2013_lr54, Suzuki2012_lr60, Chevalier2010_lr65, Lin2020_lr69, Sydow2017_lr72, Suzuki2013_lr79, Suzuki2015_lr82, Sciascio2017_lr83, Betancourt2016_lr95, Khairova2017_lr96, Ferretti2018_lr100, Dalip2011_lr111, Suzuki2012_lr117, Fahimnia2022_lr118, Hanada2013_lr125, Soonthornphisaj2017_lr130, Suzuki2013_lr133, Lewoniewski2017_lr139, Yahya2014_lr148, Lewoniewski2018_lr149, Saengthongpattana2018_lr150, Suzuki2012_lr159, Urquiza2016_lr160, Saengthongpattana2017_lr162, Saengthongpattana2014_lr169, Himoro2013_lr199, Stein2007_lr1024, Seyedsadr2016_lr2005, Yahya2020_lr2011, Olcer2022_lr2017, Wohner2015_lr2022, Xiao2013_lr2030}.

\begin{figure}[htbp]
    \centering
    \includegraphics[width=0.8\textwidth]{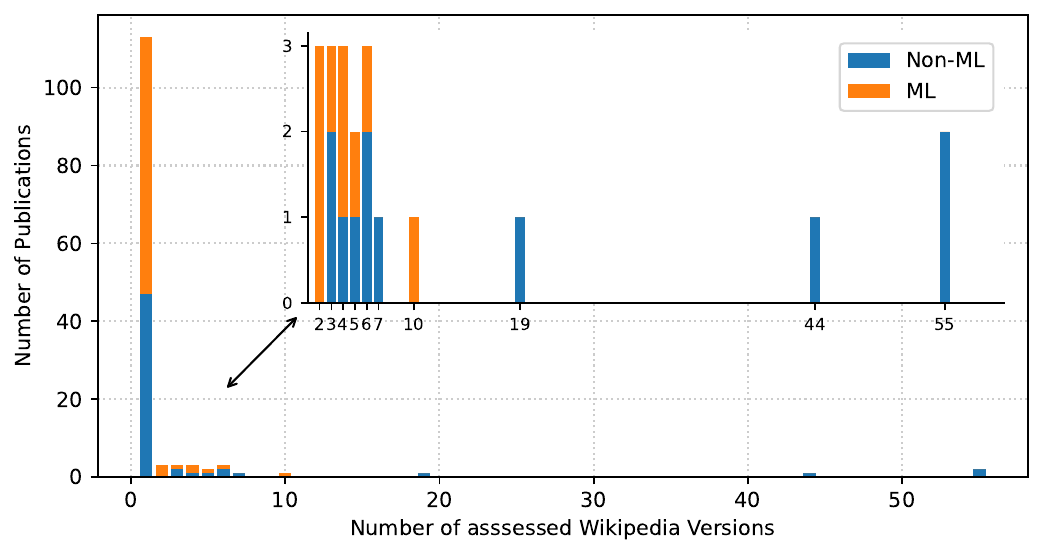}
    \caption{Number of assessed Wikipedia Versions per publication}
    \label{fig:languages_count}
\end{figure}

We also analyzed how frequently authors study multiple versions at the same time. Figure~\ref{fig:languages_count} shows that papers almost never evaluate the quality of more than one language, but one of them~\cite{Halfaker2020_lr1055} does a great job exploring this topic, designing different quality models for ten Wikipedia versions.

\subsection{Year of Publication}

\begin{figure}[htbp]
    \centering
    \includegraphics[width=0.75\textwidth]{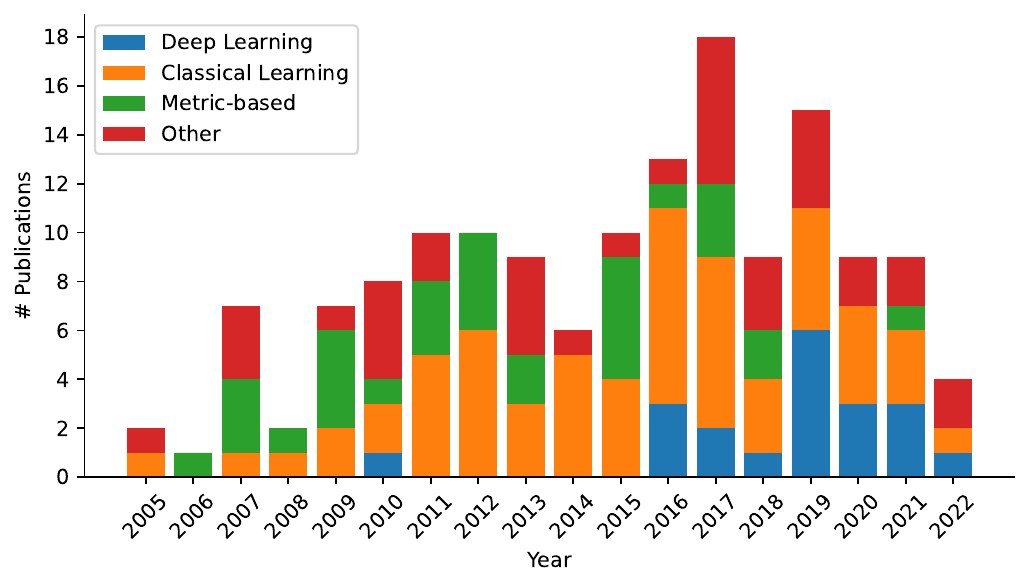}
    \caption{Included papers by year of publication}
    \label{fig:literature_trends}
\end{figure}

We analyzed the publication year of our results to study trends in this topic. Stvilia et al.'s~\cite{Stvilia2005_lr12, Stvilia2005_lr1013} studies, from 2005, were the oldest of the \PUBTOTAL, after which interest started to grow steadily. Figure~\ref{fig:literature_trends} shows that classical learning methods remained common through the years, but deep learning is clearly becoming a more prevalent approach, while metric-based studies are becoming more scarce.

\subsection{Publication Venues}

Most of the analyzed publications were published at international conferences, but we still counted many journal papers. To obtain a better overview of which publication venues are more frequent, we aggregated that information in Table~\ref{tab:venues}, which shows the conferences and journals that published more than one of the papers we included in the review.

\input{sections/results/overview/venues}

Notably, OpenSym~\footnote{OpenSym Website URL: \url{https://opensym.org/about-us/}} (formerly WikiSym) is the venue that has the most publications related to our research topic. That observation is not surprising considering their significant dedication to open collaboration research. Similarly, JASIST~\footnote{JASIST Website URL: \url{https://asistdl.onlinelibrary.wiley.com/journal/23301643}} is the peer-reviewed journal that publishes most articles on this topic.

\subsection{Publication Influence - References and Citations}

\begin{figure}[htbp]
    \centering
    \includegraphics[width=0.75\textwidth]{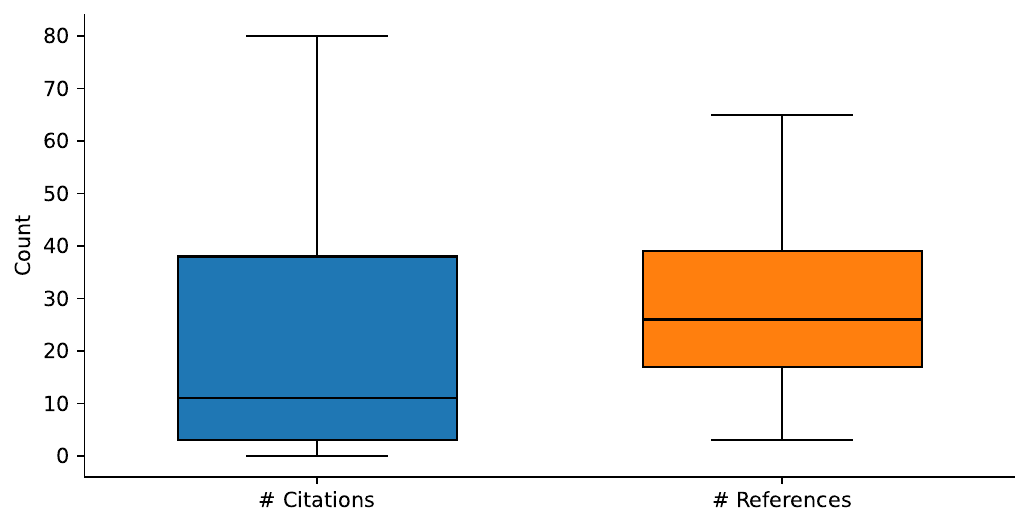}
    \caption{Overview of citation (obtained from Google Scholar in March of 2023) and reference count of included publications }
    \label{fig:num_citations}
\end{figure}

We examined and compared the number of citations and references of each included record, aiming to discover which papers were the most influential and which ones cover more sources. As shown in Figure~\ref{fig:num_citations}, citation count varies significantly across the literature, but there are still many highly cited papers. The reference count is more stable, generally between 15 and 40. For legibility purposes, we excluded outliers (fliers) from the box plot, but we still find their analysis relevant. We found several highly cited papers, such as Stvilia et al.'s ~\cite{Stvilia2007_lr1012, Stvilia2005_lr1013}, Wilkinson and Huberman's~\cite{Wilkinson2007_lr2}, Blumenstock's~\cite{Blumenstock2008_lr4}, and Hu et al.'s~\cite{Hu2007_lr1}, all of which collect over 300 citations each. The publication with the most references is Halfaker and Geiger's~\cite{Halfaker2020_lr1055}, referencing 113 other papers.

We can obtain additional conclusions from Table~\ref{tab:impactful}, which shows that deep learning methods are not as influential as the others. However, as we have seen, these solutions are just starting to emerge, so it is possible this observation changes in the future.

\input{sections/results/overview/impactful}
 
\subsection{Abstract and Keywords Analysis} 

We also analyzed the most common terms in the abstract and keywords of the included publications. To do this, we first performed text normalization~\footnote{We used the NTLK library (\url{https://www.nltk.org/}) to assist in the normalization steps.}, which included tokenization, conversion to lowercase, removal of stop words and punctuation, and simplification of all words to their singular form. Next, for each group (abstract or keywords), we computed two measures: (1) the number of times each term appears in the collection of abstracts or keywords; (2) the number of abstracts or keywords in which each term appears in. These concepts are, respectively, the collection frequency ($cf_t$) and document frequency ($df_t$) as coined in the Information Retrieval area~\cite{Manning2008_IR}. Table~\ref{tab:terms} 
summarizes this information, but most results are unsurprising. Aside from the obvious terms (e.g., quality, wikipedia, article), we can see that terms related to machine learning, edits, and network analysis frequently occur.

\input{sections/results/overview/terms}

\subsection{Authors and Affiliations}

We also decided to measure author presence across this research topic to determine which researchers study this subject more often. Table~\ref{tab:authors} summarizes this information, displaying all the authors from which we collected four or more publications, sorted by their influence, which is measured by averaging the number of citations per year of each paper we included.

\input{sections/results/overview/authors}


%

\subsection{Datasets, Source code, and External Tools}

Regardless of the followed approach, authors generally create their datasets from Wikimedia dumps~\footnote{Dumps are available through \url{https://dumps.wikimedia.org/}}, selecting a subset of articles with varying quality distributions. Unfortunately, only 18 papers publish the datasets they use~\cite{Hu2007_lr1, Javanmardi2010_lr11, Dang2016_lr16, Halfaker2017_lr22, Shen2017_lr31, Zhang2018_lr41, Lerner2018_lr57, Bassani2019_lr67, Schmidt2019_lr78, Ferretti2018_lr100, Ferretti2017_lr132, Couto2021_lr145, Pereyra2019_lr147, Lee2013_lr202, Dalip2014_lr1004, Pohn2014_lr1040, Dalip2012_lr2014, Olcer2022_lr2017}, and most of the ones we encountered were inaccessible. In terms of implementation details, 10 papers provide the source code of their study~\cite{Dang2017_lr23, Dang2016_lr24, Asthana2021_lr76, Joorabchi2019_lr84, Sarkar2019_lr87, Couto2021_lr145, Halfaker2020_lr1055, Shen2019_lr1061, Shen2020_lr2009, Olcer2022_lr2017}.

\begin{figure}[htbp]
    \centering
    \includegraphics[width=0.7\textwidth]{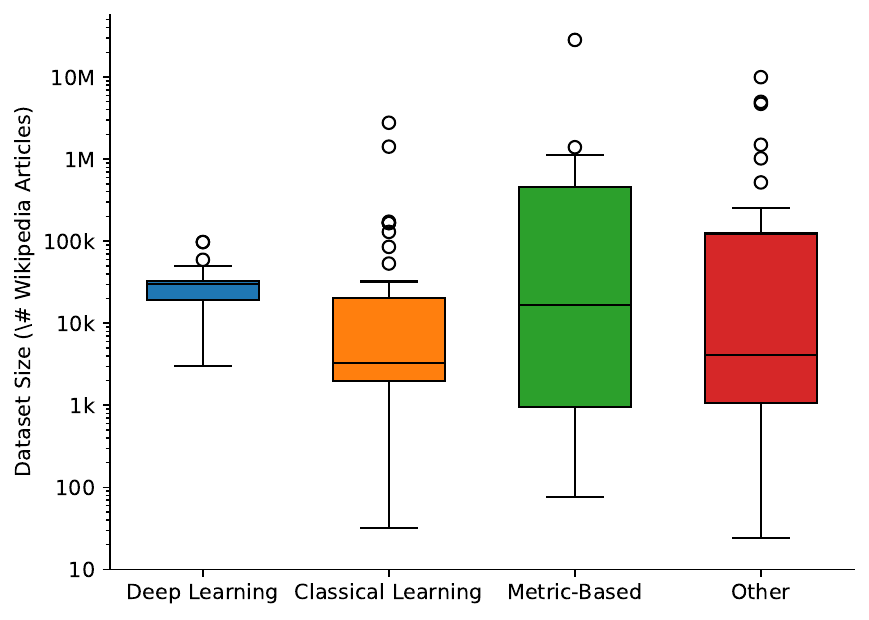}
    \caption{Size of used datasets per method type}
    \label{fig:datasets}
\end{figure}

We also analyzed the used datasets to understand how they differ between studies. Figure~\ref{fig:datasets} shows that machine learning datasets usually do not reach sizes as large as metric-based and other methods (e.g. feature-quality correlation approaches) do. This observation makes sense: training models is computationally expensive, so studies that assess Wikipedia quality without artificial intelligence can afford to use larger datasets. 

Finally, we analyzed the tools and libraries that authors most use in their studies. Table~\ref{tab:tools} shows some of the ones we collected when analyzing the manuscripts, which we hope will be useful for helping future researchers choose between technologies.

\begin{table}[htbp]
    \centering
    \caption{Relevant Tools and Libraries}
    \label{tab:tools}
    \begin{tabular}{c c l}
    \hline
    \textbf{Type} & \textbf{Name} & \textbf{URL} \\
    \hline
    \multirow{3}{*}{NLP} 
        & Doc2Vec & \url{https://radimrehurek.com/gensim/models/doc2vec.html} \\
        & Diction & \url{https://www.gnu.org/software/diction/} \\
        & Snowball & \url{https://snowballstem.org/} \\
    \hline
    \multirow{2}{*}{Classical Learning} 
        & Scikit-Learn & \url{https://scikit-learn.org/} \\
        & Weka & \url{https://waikato.github.io/weka-site/index.html} \\
    \hline
    \multirow{2}{*}{Deep Learning} 
        & Keras & \url{https://keras.io/} \\
        & TensorFlow & \url{https://www.tensorflow.org/} \\
    \hline
    \multirow{2}{*}{Wikipedia} 
        & MWParserFromHell & \url{https://mwparserfromhell.readthedocs.io/en/latest/} \\
        & WebGraph & \url{https://webgraph.di.unimi.it/} \\
    \hline
    \end{tabular}
\end{table}

%% file: sections/results/allcits.tex
\cite{Hu2007_lr1, Wilkinson2007_lr2, Liu2011_lr3, Blumenstock2008_lr4, Arazy2010_lr6, Calzada2010_lr8, Wohner2009_lr10, Javanmardi2010_lr11, Stvilia2005_lr12, Warncke-Wang2013_lr13, Dalip2009_lr14, Druck2008_lr15, Dang2016_lr16, Anderka2012_lr17, Lewoniewski2016_lr18, Halfaker2009_lr19, Li2015_lr20, Halfaker2017_lr22, Dang2017_lr23, Dang2016_lr24, Ruprechter2020_lr25, Wang2020_lr26, Liu2018_lr29, Xu2011_lr30, Shen2017_lr31, Robertie2015_lr32, Cusinato2009_lr33, Wecel2015_lr34, Anderka2011_lr35, Flekova2014_lr36, Guda2020_lr38, Zhang2018_lr41, Ferschke2012_lr43, Lewoniewski2017_lr46, Wang2021_lr48, Ingawale2013_lr54, Lerner2018_lr57, Hu2007_lr58, Dondio2007_lr59, Suzuki2012_lr60, Wu2010_lr61, Lewoniewski2018_lr62, Raman2020_lr64, Chevalier2010_lr65, Lewoniewski2019_lr66, Bassani2019_lr67, Lin2020_lr69, Wang2010_lr70, Lewoniewski2018_lr71, Sydow2017_lr72, Wang2019_lr74, Asthana2021_lr76, Schmidt2019_lr78, Suzuki2013_lr79, Suzuki2015_lr82, Sciascio2017_lr83, Joorabchi2019_lr84, Marrese-Taylor2019_lr85, Sarkar2019_lr87, Dang2016_lr89, Moturu2009_lr90, 
Cozza2016_lr92, Betancourt2016_lr95, Khairova2017_lr96, Das2021_lr97, Ferretti2018_lr100, Chhabra2021_lr101, Lewoniewski2017_lr106, Lewoniewski2017_lr109, Dalip2011_lr111, Khan2019_lr114, Ferretti2012_lr115, Suzuki2012_lr117, Fahimnia2022_lr118, Sugandhika2022_lr119, Hou2021_lr122, Hanada2013_lr125, Ruprechter2019_lr127, Su2015_lr128, Soonthornphisaj2017_lr130, Qin2012_lr131, Ferretti2017_lr132, Suzuki2013_lr133, Dang2021_lr136, Lewoniewski2017_lr139, 
Velichety2019_lr142, Couto2021_lr145, Pereyra2019_lr147, Yahya2014_lr148, Lewoniewski2018_lr149, Saengthongpattana2018_lr150, Tsikerdekis2017_lr155, Kleminski2017_lr157, Suzuki2012_lr159, Urquiza2016_lr160, Couto2021_lr161, Saengthongpattana2017_lr162, Saengthongpattana2014_lr169, Robertie2017_lr172, Lu2013_lr173, Deng2015_lr181, Zhang2015_lr197, Himoro2013_lr199, Lee2013_lr202, Jemielniak2017_lr262, Velazquez2017_lr338, Bassani2019_lr359, Agrawal2016_lr1001, Dalip2016_lr1002, Dalip2011_lr1003, Dalip2014_lr1004, Ganjisaffar2009_lr1006, Ofek2015_lr1010, Stvilia2007_lr1012, Stvilia2005_lr1013, Ruvo2015_lr1016, Lipka2010_lr1019, Rassbach2007_lr1020, Wu2012_lr1021, Stein2007_lr1024, Lex2012_lr1026, Anderka2011_lr1027, Wu2011_lr1030, Stvilia2009_lr1038, Pohn2014_lr1040, Sugandhika2021_lr1041, Lim2006_lr1044, Hammwohner2010_lr1046, Han2011_lr1051, Halfaker2020_lr1055, Shen2019_lr1061, Yang2016_lr2001, Velichety2019_lr2002, Seyedsadr2016_lr2005, Ge2020_lr2008, Shen2020_lr2009, Marzini2014_lr2010, Yahya2020_lr2011, Hu2016_lr2013, Dalip2012_lr2014, Han2011_lr2015, Han2011_lr2016, Olcer2022_lr2017, Li2022_lr2019, Wohner2015_lr2022, Yu2018_lr2024, Magalhaes2019_lr2028, Gopalan2016_lr2029, Xiao2013_lr2030},

%% file: sections/results/overview/methods.tex
\begin{table}[htbp]
    \caption{Most popular approaches}
    \label{tab:methods}
    \centering
    \begin{tabular}{c c m{0.65\textwidth}}
        \toprule
        \textbf{Approach} & \textbf{\# Papers} & \textbf{ }\\
        \midrule
        CL + Features & 51 & \cite{Blumenstock2008_lr4, Warncke-Wang2013_lr13, Dalip2009_lr14, Dang2016_lr16, Anderka2012_lr17, Lewoniewski2016_lr18, Halfaker2017_lr22, Ruprechter2020_lr25, Xu2011_lr30, Wecel2015_lr34, Anderka2011_lr35, Flekova2014_lr36, Ferschke2012_lr43, Raman2020_lr64, Bassani2019_lr67, Wang2010_lr70, Cozza2016_lr92, Ferretti2018_lr100, Ferretti2012_lr115, Su2015_lr128, Soonthornphisaj2017_lr130, Ferretti2017_lr132, Dang2021_lr136, Velichety2019_lr142, Yahya2014_lr148, Saengthongpattana2018_lr150, Urquiza2016_lr160, Saengthongpattana2014_lr169, Robertie2017_lr172, Lu2013_lr173, Zhang2015_lr197, Bassani2019_lr359, Dalip2016_lr1002, Dalip2011_lr1003, Dalip2014_lr1004, Ganjisaffar2009_lr1006, Ofek2015_lr1010, Stvilia2005_lr1013, Lipka2010_lr1019, Rassbach2007_lr1020, Lex2012_lr1026, Pohn2014_lr1040, Sugandhika2021_lr1041, Halfaker2020_lr1055, Velichety2019_lr2002, Seyedsadr2016_lr2005, Yahya2020_lr2011, Dalip2012_lr2014, Magalhaes2019_lr2028, Gopalan2016_lr2029, Xiao2013_lr2030} \\
        Deep Learning & 20 & \cite{Dang2017_lr23, Dang2016_lr24, Wang2020_lr26, Shen2017_lr31, Guda2020_lr38, Zhang2018_lr41, Wang2021_lr48, Wu2010_lr61, Wang2019_lr74, Asthana2021_lr76, Schmidt2019_lr78, Marrese-Taylor2019_lr85, Sarkar2019_lr87, Dang2016_lr89, Hou2021_lr122, Pereyra2019_lr147, Agrawal2016_lr1001, Shen2019_lr1061, Shen2020_lr2009, Li2022_lr2019} \\
        Metric-based & 31 & \cite{Hu2007_lr1, Wohner2009_lr10, Javanmardi2010_lr11, Druck2008_lr15, Halfaker2009_lr19, Li2015_lr20, Robertie2015_lr32, Cusinato2009_lr33, Lewoniewski2017_lr46, Hu2007_lr58, Suzuki2012_lr60, Lewoniewski2018_lr71, Suzuki2013_lr79, Suzuki2015_lr82, Moturu2009_lr90, Suzuki2012_lr117, Qin2012_lr131, Suzuki2013_lr133, Couto2021_lr145, Kleminski2017_lr157, Suzuki2012_lr159, Deng2015_lr181, Velazquez2017_lr338, Ruvo2015_lr1016, Stein2007_lr1024, Lim2006_lr1044, Han2011_lr1051, Hu2016_lr2013, Han2011_lr2015, Han2011_lr2016, Yu2018_lr2024} \\
        Feat. Correlation & 20 & \cite{Calzada2010_lr8, Lewoniewski2017_lr46, Ingawale2013_lr54, Lerner2018_lr57, Lewoniewski2019_lr66, Lin2020_lr69, Joorabchi2019_lr84, Khairova2017_lr96, Das2021_lr97, Lewoniewski2017_lr106, Lewoniewski2017_lr109, Khan2019_lr114, Ruprechter2019_lr127, Yahya2014_lr148, Himoro2013_lr199, Jemielniak2017_lr262, Stvilia2009_lr1038, Hammwohner2010_lr1046, Ge2020_lr2008, Marzini2014_lr2010} \\
        \bottomrule
    \end{tabular}
\end{table}

%% file: sections/results/overview/venues.tex
\begin{table}[htbp]
    \caption{Overview of the venues}
    \label{tab:venues}
    \centering
    \begin{tabular}{c m{0.65\textwidth} c}
        \toprule
        \textbf{Type} & \textbf{Venue} & \textbf{\# Papers} \\
        \midrule
        Conference & OpenSym$^*$: International Symposium on Open Collaboration & 14 \\
        Conference & WWW: The Web Conference & 6 \\
        Conference & BIS: International Conference on Business Information Systems & 6 \\
        Conference & JCDL: ACM/IEEE Joint Conference on Digital Libraries & 5 \\
        Journal & Journal of the Association for Information Science and Technology & 5 \\
        Conference & ICIST: International Conference on Information and Software Technologies & 4 \\
        Conference & CIKM: International Conference on Information and Knowledge Management & 3 \\
        Journal & Proceedings of the ACM on Human-Computer Interaction & 3 \\
        Conference & CLEF: Conference and Labs of the Evaluation Forum & 2 \\
        Journal & Expert Systems with Applications & 2 \\
        Journal & Online Information Review & 2 \\
        Journal & Journal of Information Processing & 2 \\
        Conference & WorldCIST: World Conference on Information Systems and Technologies & 2 \\
        Conference & HT: ACM Conference on Hypertext \& Social Media & 2 \\
        Conference & WebMedia: Brazilian Symposium on Multimedia and the Web & 2 \\
        Conference & CACIC: Argentine Congress of Computer Science & 2 \\
        Conference & iSAI-NLP: International Joint Symposium on Artificial Intelligence and Natural Language Processing & 2 \\
        Conference & WAIM: International Conference on Web-Age Information Management & 2 \\
        Conference & WI: IEEE WIC ACM International Conference on Web Intelligence & 2 \\
        \bottomrule
    \end{tabular}
    \\ \vspace{0.1cm}
    \footnotesize
    $^*$ Formerly WikiSym
\end{table}

%% file: sections/results/overview/impactful.tex
\begin{table}[htbp]
    \caption{Top 15 most impactful publications}
    \label{tab:impactful}
    \centering
    \begin{tabular}{m{0.69\textwidth} c c}
        \toprule
        \textbf{Study} & \textbf{Type}$^{(1)}$ & \textbf{Impact}$^{(2)}$ \\
        \midrule
        A framework for information quality assessment~\cite{Stvilia2007_lr1012} & FMC & 37.88 / 13.12 / 19.19 \\
        Size matters: word count as a measure of quality on wikipedia~\cite{Blumenstock2008_lr4} & CL & 25.27 / N/A / 14.53 \\
        Cooperation and quality in wikipedia~\cite{Wilkinson2007_lr2} & FMC & 24.38 / N/A / 10.12 \\
        ORES: Lowering Barriers with Participatory Machine Learning in Wikipedia~\cite{Halfaker2020_lr1055} & CL & 23.33 / N/A / 7.67 \\
        Measuring article quality in wikipedia: models and evaluation~\cite{Hu2007_lr1} & MB & 22.5 / N/A / 11.38 \\
        Assessing information quality of a community-based encyclopedia~\cite{Stvilia2005_lr1013} & CL & 22.28 / N/A / 11.5 \\
        Who does what: Collaboration patterns in the wikipedia and their impact on article quality~\cite{Liu2011_lr3} & FMC & 14.33 / N/A / 8.58 \\
        Tell me more: an actionable quality model for Wikipedia~\cite{Warncke-Wang2013_lr13} & CL & 12.6 / N/A / 8.1 \\
        NwQM: A Neural Quality Assessment Framework for Wikipedia~\cite{Guda2020_lr38} & DL & 12.33 / 1.0 / 1.67 \\
        Automatic quality assessment of content created collaboratively by web communities: a case study of wikipedia~\cite{Dalip2009_lr14} & CL & 10.71 / 3.29 / 6.36 \\
        Assessing the quality of information on wikipedia: A deep‐learning approach~\cite{Wang2020_lr26} & DL & 10.67 / 3.67 / 5.67 \\
        Assessing the quality of Wikipedia articles with lifecycle based metrics~\cite{Wohner2009_lr10} & MB & 10.0 / N/A / 5.43 \\
        Predicting quality flaws in user-generated content: the case of wikipedia~\cite{Anderka2012_lr17} & CL & 9.91 / 2.82 / 5.18 \\
        Who Did What: Editor Role Identification in Wikipedia~\cite{Yang2016_lr2001} & FMC & 9.57 / N/A / 5.86 \\
        Information quality discussions in wikipedia~\cite{Stvilia2005_lr12} & FMC & 9.33 / N/A / N/A \\
        \bottomrule
    \end{tabular}
    \vspace{0.1cm}
    \footnotesize
    \RaggedRight
    \\$^{(1)}$ CL = Classical Learning, DL = Deep Learning, MB = Metric-based, FMC = Feature/Metric quality correlation
    \\$^{(2)}$ Measured as the number of citations per year of publication. Citation data obtained from Google Scholar, Web of Science, and Scopus, respectively, in March of 2023. N/A indicates the publication was not found in the respective database.
\end{table}

%% file: sections/results/overview/terms.tex
\begin{table}[ht]
    \caption{Term Analysis: Ten highest Collection and Document Frequency of Abstract ($cf_{ta}$, $df_{ta}$) and Keywords ($cf_{tk}$, $df_{tk}$) terms}
    \label{tab:terms}
    \begin{minipage}{.2\textwidth}
        \centering
        \begin{tabular}{c c}
            \toprule
            Term ($t$) & $cf_{ta}$ \\
            \midrule
            quality & 652 \\
            article & 594 \\
            wikipedia & 472 \\
            model & 151 \\
            feature & 146 \\
            content & 133 \\
            information & 117 \\
            approach & 98 \\
            editor & 85 \\
            assessment & 85 \\
            \bottomrule
        \end{tabular}
    \end{minipage}
    \begin{minipage}{.2\textwidth}
        \centering
        \begin{tabular}{c c}
            \toprule
            Term ($t$) & $df_{ta}$ \\
            \midrule
            quality & 146 \\
            wikipedia & 139 \\
            article & 136 \\
            content & 73 \\
            paper & 72 \\
            result & 65 \\
            information & 63 \\
            model & 58 \\
            approach & 54 \\
            feature & 54 \\
            \bottomrule
        \end{tabular}
    \end{minipage}
    \begin{minipage}{.2\textwidth}
        \centering
        \begin{tabular}{c c}
            \toprule
            Term ($t$) & $cf_{tk}$ \\
            \midrule
            quality & 116 \\
            wikipedia & 106 \\
            information & 34 \\
            article & 32 \\
            learning & 22 \\
            assessment & 21 \\
            data & 14 \\
            analysis & 13 \\
            network & 13 \\
            edit & 13 \\
            \bottomrule
        \end{tabular}
    \end{minipage}
    \begin{minipage}{.2\textwidth}
        \centering
        \begin{tabular}{c c}
            \toprule
            Term ($t$) & $df_{tk}$ \\
            \midrule
            wikipedia & 104 \\
            quality & 96 \\
            information & 30 \\
            article & 29 \\
            assessment & 20 \\
            learning & 20 \\
            analysis & 13 \\
            network & 13 \\
            classification & 12 \\
            machine & 12 \\
            \bottomrule
        \end{tabular}
    \end{minipage}
\end{table}

%% file: sections/results/overview/authors.tex
\begin{table}[htbp]
    \caption{Authors with five or more included publications in the literature review}
    \label{tab:authors}
    \centering
    \begin{tabular}{l c l c} 
        \toprule
        \textbf{Author} & \textbf{Cits. / Year} & \textbf{Publications} & \textbf{(\#)} \\
        \midrule
        Aaron L. Halfaker & 9.73 & \cite{Halfaker2009_lr19, Halfaker2017_lr22, Asthana2021_lr76, Halfaker2020_lr1055, Yang2016_lr2001} & (5) \\
        Benno Stein & 6.1 & \cite{Anderka2012_lr17, Anderka2011_lr35, Lipka2010_lr1019, Lex2012_lr1026, Anderka2011_lr1027} & (5) \\
        Quang-Vinh Dang & 4.76 & \cite{Dang2016_lr16, Dang2017_lr23, Dang2016_lr24, Dang2016_lr89, Dang2021_lr136} & (5) \\
        Pável Calado & 4.37 & \cite{Dalip2009_lr14, Dalip2016_lr1002, Dalip2011_lr1003, Dalip2014_lr1004, Dalip2012_lr2014, Magalhaes2019_lr2028} & (6) \\
        Daniel Hasan Dalip & 3.82 & \cite{Dalip2009_lr14, Dalip2011_lr111, Dalip2016_lr1002, Dalip2011_lr1003, Dalip2014_lr1004, Dalip2012_lr2014, Magalhaes2019_lr2028} & (7) \\
        Marcos André Gonçalves & 3.82 & \cite{Dalip2009_lr14, Dalip2011_lr111, Dalip2016_lr1002, Dalip2011_lr1003, Dalip2014_lr1004, Dalip2012_lr2014, Magalhaes2019_lr2028} & (7) \\
        Ping Wang & 3.63 & \cite{Wang2020_lr26, Wang2021_lr48, Wang2019_lr74, Hou2021_lr122, Li2022_lr2019} & (5) \\
        Witold Abramowicz & 3.44 & \cite{Lewoniewski2016_lr18, Lewoniewski2017_lr46, Lewoniewski2019_lr66, Lewoniewski2018_lr71, Lewoniewski2017_lr139, Lewoniewski2018_lr149} & (6) \\
        Marco Cristo & 3.39 & \cite{Dalip2009_lr14, Hanada2013_lr125, Himoro2013_lr199, Dalip2016_lr1002, Dalip2011_lr1003, Dalip2014_lr1004, Dalip2012_lr2014, Magalhaes2019_lr2028} & (8) \\
        Krzysztof Węcel & 3.12 & \cite{Lewoniewski2016_lr18, Wecel2015_lr34, Lewoniewski2017_lr46, Lewoniewski2019_lr66, Lewoniewski2018_lr71, Khairova2017_lr96, Lewoniewski2017_lr106, Lewoniewski2017_lr139, Lewoniewski2018_lr149} & (9) \\
        Włodzimierz Lewoniewski & 2.88 & \cite{Lewoniewski2016_lr18, Wecel2015_lr34, Lewoniewski2017_lr46, Lewoniewski2018_lr62, Lewoniewski2019_lr66, Lewoniewski2018_lr71, Khairova2017_lr96, Lewoniewski2017_lr106, Lewoniewski2017_lr109, Lewoniewski2017_lr139, Lewoniewski2018_lr149, Ge2020_lr2008} & (12) \\
        Yu Suzuki & 1.64 & \cite{Suzuki2012_lr60, Suzuki2013_lr79, Suzuki2015_lr82, Suzuki2012_lr117, Suzuki2013_lr133, Suzuki2012_lr159} & (6) \\
        Marcelo Errecalde & 1.63 & \cite{Sciascio2017_lr83, Ferretti2018_lr100, Ferretti2012_lr115, Ferretti2017_lr132, Pereyra2019_lr147, Urquiza2016_lr160, Velazquez2017_lr338, Lex2012_lr1026, Pohn2014_lr1040} & (9) \\
        Edgardo Ferretti & 1.62 & \cite{Ferretti2018_lr100, Ferretti2012_lr115, Ferretti2017_lr132, Pereyra2019_lr147, Urquiza2016_lr160, Lex2012_lr1026, Pohn2014_lr1040} & (7) \\
        \bottomrule
    \end{tabular}
\end{table}

%% file: sections/section5-res-ml.tex
\section{Machine Learning Approaches} \label{sec:res-ml}

Here we describe the approaches used by the \PUBML\ papers using machine learning to evaluate the quality of Wikipedia articles, comparing their performance. Authors do not always report their results using the same performance metrics, so it is not trivial to compare them directly. We wish to summarize the literature concisely but rigorously, so we will present this section's results sorted by performance value, clearly indicating the metric chosen by each study. We will not list here any of the \PUBBADMETRICS\ papers that do not use Accuracy, ROC AUC, or F1-score, but we still collected those experiments for our dataset. Results here will prioritize showing AUC over accuracy, as it is often considered a better measure~\cite{Ling2003_AUCvsAcc}, and we will only show the F1-score if none of the two previous measures were found.

\begin{figure}[htbp]
    \centering
    \includegraphics[width=0.6\textwidth]{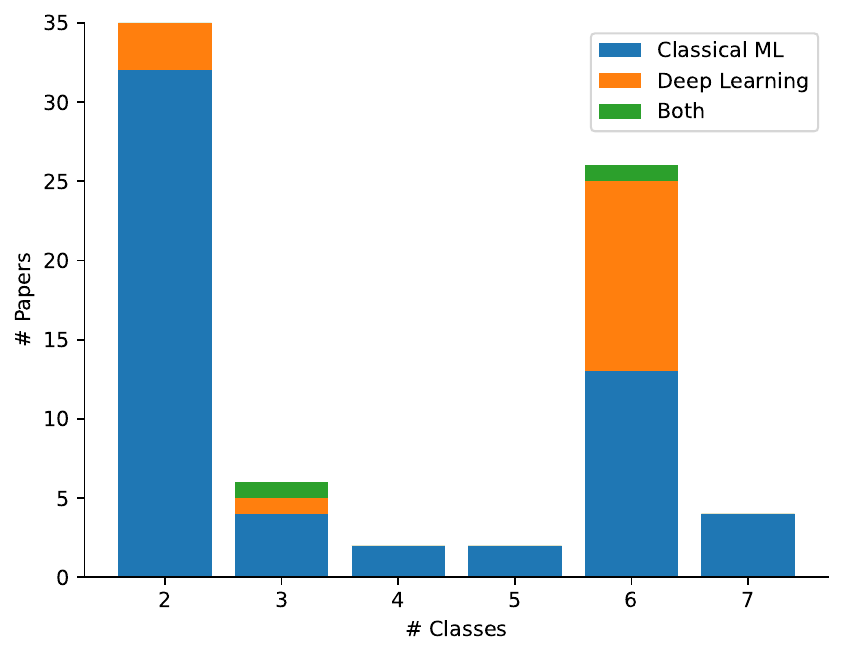}
    \caption{Number of classes considered in machine learning experiments}
    \label{fig:num_classes}
\end{figure}

As Figure~\ref{fig:num_classes} suggests, 2-class and 6-class setups are much more common than the rest, so we will mainly focus on the performance of those solutions. We will separate those to allow us to better compare study performances, but we must first pay attention to the quality labels used for each class. Most 6-class studies consider Wikipedia's Stub to FA scale~\cite{Web_WikipediaContentAss} (usually excluding A-tier), and 2-class typically follow a \textit{Featured Article vs. Random Article} approach, so those comparisons should be safe. Also, since performance varies with both the number and distribution of considered classes, for each study we also show the dataset's imbalance ratio (\textit{$IR$} = \textit{\# samples in the majority class} / \textit{\# samples in the minority class})~\cite{Zhu2020_ImbalanceRatio}. 

Due to the nature of Machine Learning algorithms, it is unlikely that the best approach will be the same for every dataset. In fact, the \emph{No Free Lunch} Theorem~\cite{Wolpert1997_NoFreeLunch} states that all optimization algorithms have the same performance when averaged across all possible problems. Regardless, we collected all the results to understand better which algorithms were experimented with, and how performant they are in the given conditions, providing a baseline for future studies.

\subsection{Classical Learning}

We have seen that classical learning algorithms are the most common methods in this review: \PUBCL\ publications opt to use them~\cite{Blumenstock2008_lr4, Warncke-Wang2013_lr13, Dalip2009_lr14, Dang2016_lr16, Anderka2012_lr17, Lewoniewski2016_lr18, Halfaker2017_lr22, Ruprechter2020_lr25, Xu2011_lr30, Wecel2015_lr34, Anderka2011_lr35, Flekova2014_lr36, Ferschke2012_lr43, Wang2021_lr48, Raman2020_lr64, Bassani2019_lr67, Wang2010_lr70, Sydow2017_lr72, Wang2019_lr74, Schmidt2019_lr78, Cozza2016_lr92, Betancourt2016_lr95, Ferretti2018_lr100, Chhabra2021_lr101, Ferretti2012_lr115, Sugandhika2022_lr119, Su2015_lr128, Soonthornphisaj2017_lr130, Ferretti2017_lr132, Dang2021_lr136, Lewoniewski2017_lr139, Velichety2019_lr142, Pereyra2019_lr147, Yahya2014_lr148, Lewoniewski2018_lr149, Saengthongpattana2018_lr150, Urquiza2016_lr160, Saengthongpattana2017_lr162, Saengthongpattana2014_lr169, Robertie2017_lr172, Lu2013_lr173, Zhang2015_lr197, Bassani2019_lr359, Dalip2016_lr1002, Dalip2011_lr1003, Dalip2014_lr1004, Ganjisaffar2009_lr1006, Ofek2015_lr1010, Stvilia2005_lr1013, Lipka2010_lr1019, Rassbach2007_lr1020, Wu2012_lr1021, Lex2012_lr1026, Anderka2011_lr1027, Wu2011_lr1030, Pohn2014_lr1040, Sugandhika2021_lr1041, Halfaker2020_lr1055, Velichety2019_lr2002, Seyedsadr2016_lr2005, Yahya2020_lr2011, Dalip2012_lr2014, Magalhaes2019_lr2028, Gopalan2016_lr2029, Xiao2013_lr2030}. Tables~\ref{tab:CL_performance_6class} and~\ref{tab:CL_performance_2class} show that decision trees, random forests, and SVMs are frequently great classical approaches, but the used performance metrics and class distribution vary so much that it is difficult to determine which solution is best. We also noticed that the best methods are almost always trained on English data, and those that are trained on multiple languages typically show much worse results on non-English data (e.g., Halfaker and Geiger~\cite{Halfaker2020_lr1055}), which suggests there is a need for more work on multilingual assessment.

\input{sections/results/ml/performance.CL.6class}

\input{sections/results/ml/performance.CL.2class}

Although quality might seem a continuous measure, almost all authors decided to solve a classification task. However, wrong predictions are typically not far from the correct ones. In fact, papers sometimes present off-by-one-class accuracy in their results~\cite{Halfaker2017_lr22}, which tend to be much higher. Only eight studies~\cite{Dalip2009_lr14, Dang2021_lr136, Dalip2016_lr1002, Dalip2011_lr1003, Dalip2014_lr1004, Seyedsadr2016_lr2005, Dalip2012_lr2014, Magalhaes2019_lr2028} tackle this problem as a regression task, but the method of solving it is very similar to others, typically using a feature-based approach. 

\subsection{Deep Learning}

Although less common, deep learning methods have recently been gaining more relevance in this field. Among the \PUBTOTAL\ publications we collected, \PUBDL\ of them use deep learning~\cite{Dang2017_lr23, Dang2016_lr24, Wang2020_lr26, Shen2017_lr31, Guda2020_lr38, Zhang2018_lr41, Wang2021_lr48, Wu2010_lr61, Wang2019_lr74, Asthana2021_lr76, Schmidt2019_lr78, Marrese-Taylor2019_lr85, Sarkar2019_lr87, Dang2016_lr89, Hou2021_lr122, Pereyra2019_lr147, Agrawal2016_lr1001, Shen2019_lr1061, Shen2020_lr2009, Li2022_lr2019}. Tables~\ref{tab:DL_performance_6class} and~\ref{tab:DL_performance_2class} suggest that LSTMs and GRUs lead to the most promising results, often better than classical methods. Unfortunately, we could not collect class distribution information from many studies, which makes us uncertain about how to best assess these results. Once again, we notice a strong preference for English datasets over non-English ones.

\input{sections/results/ml/performance.DL.6class}

\input{sections/results/ml/performance.DL.2class}

%% file: sections/results/ml/performance.CL.6class.tex
\begin{table}[ht]
    \caption{Classical Learning accuracy of 6-class approaches }
    \label{tab:CL_performance_6class}
    \centering
    \begin{tabular}{m{.3\textwidth} l l r r l}
        \toprule
        \textbf{Study} & \textbf{Best Method} & \textbf{Measure} & \textbf{Performance} & \textbf{IR$^*$} & \textbf{Lang.} \\ 
        \midrule
        Włodzimierz et al.~\cite{Lewoniewski2016_lr18} & Random Forest & AUC & 0.90 & 1.00 & Multiple \\
        Vittoria et al.~\cite{Cozza2016_lr92} & Random Forest & AUC & 0.89 & 95.06 & English \\
        Schmidt and Zangerle~\cite{Schmidt2019_lr78} & Gradient Boosted Trees & Accuracy & 73.00\% & 1.10 & English \\
        Dang and Ignat~\cite{Dang2016_lr16} & Random Forest & Accuracy & 64.00\% & 1.77 & English \\
        Halfaker~\cite{Halfaker2017_lr22} & ORES (Gradient Boosting) & Accuracy & 62.90\% & 1.12 & English \\
        Halfaker and Geiger~\cite{Halfaker2020_lr1055} & Gradient Boosting & Accuracy & 62.90\% & 1.10 & Multiple \\
        Narun et al.~\cite{Raman2020_lr64} & MLR & Accuracy & 49.35\% & 1.00 & English \\
        \bottomrule
    \end{tabular}
    \\ \vspace{0.1cm}
    \footnotesize
    $^*$ Imbalance Ratio ($IR$) = \# samples in the majority class / \# samples in the minority class. 
\end{table}

%% file: sections/results/ml/performance.CL.2class.tex
\begin{table}[ht]
    \caption{Classical Learning accuracy of 2-class approaches (Top 10)}
    \label{tab:CL_performance_2class}
    \centering
    \begin{tabular}{m{.3\textwidth} l l r r l}
        \toprule
        \textbf{Study} & \textbf{Best Method} & \textbf{Measure} & \textbf{Performance} & \textbf{IR$^*$} & \textbf{Lang.} \\ 
        \midrule
        Saengthongpattana and Soonthornphisaj~\cite{Saengthongpattana2014_lr169} & Naive Bayes & AUC & 0.99 & 236.00 & Thai \\
        Blumenstock~\cite{Blumenstock2008_lr4} & MLP & Accuracy & 97.15\% & 6.12 & English \\
        Ofek and Rokach~\cite{Ofek2015_lr1010} & Bayes Network & AUC & 0.97 & 1.00 & English \\
        Sugandhika and Ahangama~\cite{Sugandhika2022_lr119} & Logistic Regression & Accuracy & 96.00\% & 1.00 & English \\
        Adnan et al.~\cite{Yahya2020_lr2011} & Random Forest & Accuracy & 95.50\% & 1.01 & Arabic \\
        Kui et al.~\cite{Xiao2013_lr2030} & C4.5 Decision Tree & Accuracy & 94.60\% & 3.00 & Chinese \\
        Maik et al.~\cite{Anderka2011_lr35} & Random Forest & AUC & 0.94 & 1.00 & English \\
        Besiki et al.~\cite{Stvilia2005_lr1013} & C4.5 Decision Tree & F1-score & 0.94 & 3.51 & English \\
        Lipka and Stein~\cite{Lipka2010_lr1019} & SVM & Accuracy & 94.00\% & 1.00 & English \\
        Lian et al.~\cite{Pohn2014_lr1040} & SVM & F1-score & 0.94 & 1.00 & English \\
        \bottomrule
    \end{tabular}
    \\ \vspace{0.1cm}
    \footnotesize
    $^*$ Imbalance Ratio ($IR$) = \# samples in the majority class / \# samples in the minority class. 
\end{table}

%% file: sections/results/ml/performance.DL.6class.tex
\begin{table}[ht]
    \caption{Deep Learning accuracy of 6-class approaches }
    \label{tab:DL_performance_6class}
    \centering
    \begin{tabular}{m{.3\textwidth} l l r r l}
        \toprule
        \textbf{Study} & \textbf{Best Method} & \textbf{Measure} & \textbf{Performance} & \textbf{IR$^*$} & \textbf{Lang.} \\ 
        \midrule
        Jingrui et al.~\cite{Hou2021_lr122} & Stacked Learning & Accuracy & 75.46\% & ? & English \\
        Shiyue et al.~\cite{Zhang2018_lr41} & RNN + LSTM & Accuracy & 68.60\% & 1.16 & English \\
        Aili et al.~\cite{Shen2017_lr31} & Bi-LSTM+ & Accuracy & 68.17\% & 1.00 & English \\
        Dang and Ignat~\cite{Dang2017_lr23} & RNN + LSTM & Accuracy & 68.00\% & 1.01 & Multiple \\
        Edison et al.~\cite{Marrese-Taylor2019_lr85} & Bi-LSTM & Accuracy & 66.56\% & ? & Multiple \\
        Bhanu et al.~\cite{Guda2020_lr38} & BERT + GRU & Accuracy & 63.23\% & 1.64 & English \\
        Aili et al.~\cite{Shen2020_lr2009} & BiLSTM & Accuracy & 62.50\% & 1.02 & English \\
        Aili et al.~\cite{Shen2019_lr1061} & BiLSTM & Accuracy & 59.40\% & ? & English \\
        Dang and Ignat~\cite{Dang2016_lr24} & DNN & Accuracy & 55.00\% & ? & English \\
        Dang and Ignat~\cite{Dang2016_lr89} & DNN & Accuracy & 55.00\% & ? & English \\
        \bottomrule
    \end{tabular}
    \\ \vspace{0.1cm}
    \footnotesize
    $^*$ Imbalance Ratio ($IR$) = \# samples in the majority class / \# samples in the minority class. '?' indicates that we could not collect enough information about class distribution.
\end{table}

%% file: sections/results/ml/performance.DL.2class.tex
\begin{table}[ht]
    \caption{Deep Learning accuracy of 2-class approaches }
    \label{tab:DL_performance_2class}
    \centering
    \begin{tabular}{m{.3\textwidth} l l r r l}
        \toprule
        \textbf{Study} & \textbf{Best Method} & \textbf{Measure} & \textbf{Performance} & \textbf{IR$^*$} & \textbf{Lang.} \\ 
        \midrule
        Muyan et al.~\cite{Li2022_lr2019} & BERT + GRU & Accuracy & 97.58\% & 1.00 & Multiple \\
        Wang and Li~\cite{Wang2020_lr26} &  Stacked LSTM & Accuracy & 79.81\% & ? & English \\
        Sumit et al.~\cite{Asthana2021_lr76} & RNN & F1-score & 0.69 & 1.00 & English \\
        \bottomrule
    \end{tabular}
    \\ \vspace{0.1cm}
    \footnotesize
    $^*$ Imbalance Ratio ($IR$) = \# samples in the majority class / \# samples in the minority class. '?' indicates that we could not collect enough information about class distribution.
\end{table}

%% file: sections/section6-res-features.tex
\section{Article Features and Quality Metrics} \label{sec:res-features}

Some studies, typically deep learning ones, simply feed the article's full text to their model to obtain a quality prediction~\cite{Dang2017_lr23, Dang2016_lr24, Guda2020_lr38, Wang2021_lr48, Sarkar2019_lr87, Dang2016_lr89, Hou2021_lr122, Shen2019_lr1061, Shen2020_lr2009, Li2022_lr2019}, usually based on the Doc2Vec model~\cite{Le2014_Doc2Vec}. However, most approaches still use article features or metrics, with and without machine learning.

This section overviews the article features and metrics we identified in this literature review. The distinction between features and metrics varies within the papers, sometimes used interchangeably. Here, we consider something a metric if it is not reasonably simple to compute and is used by the authors as a direct measure of quality (e.g., PeerReview~\cite{Hu2007_lr1}). In contrast, features are more straightforward and indirect quality measures (e.g., Character Count).

\subsection{Article Features}

We assigned a unique ID to all the features we collected from the reviewed publications, and each falls within one of the following categories:
\begin{itemize}[itemsep=0pt]
  \item \textbf{Content features}, which relate to the length and structure of the article, taking into account factors such as the number of words, sections, or images.
  \item \textbf{Style features}, that measure how the authors write the articles, how long their phrases are, and what classes of words they use.
  \item \textbf{Readability features} estimate ``the age or US grade level necessary to comprehend a text. (...) good articles should be well written, understandable, and free of unnecessary complexity''~\cite{Dalip2009_lr14}, by measuring the sentence and word complexity. They are characterized by their use of straightforward formulas that combine other types of features.
  \item \textbf{History features}, which analyze the review history of an article and related factors, namely the article's age and the number of contributions.
  \item \textbf{Network features} are a bit more complex, as they take into account the connections between Wikipedia articles to measure their influence.
  \item \textbf{Popularity features} track the engagement of the page, analyzing values related to the number of views and visitors.
\end{itemize}

We based this categorization on the work of previous authors (e.g., Bassani and Viviani~\cite{Bassani2019_lr359}, Dalip et al.~\cite{Dalip2011_lr1003}), but there may be slight modifications. For instance, we consider internal link counts as content features, as we believe that any measure that can be directly computed through an article's wikitext should belong to the Content, Style, or Readability category. Besides, although it is frequent for authors to assign internal link counts to the network category, external link counts are rarely considered network features, and we preferred to maintain consistency. Additionally, authors usually consider Content, Style, and Readability to be subcategories of \emph{Text Features}. However, we distinguish them as different types in this review, aiming to reduce the disparity between the number of features per category.

Besides assigning a category, we also classify article features into two extra dimensions: \textbf{actionable} and \textbf{multilingual}. A feature is actionable if it can directly suggest how to improve the quality of the respective article, as proposed by Warncke-Wang et al.~\cite{Warncke-Wang2013_lr13}. For instance, a low character count may indicate that expanding the article is beneficial for its quality. As for features that are technically manipulable but not in a relevant manner to the overall goal (e.g. revision count), we do not consider them actionable. The multilingual dimension answers the question: ``Can this feature be applied to All, Most, or Some Wikipedia languages?'' This is a relevant aspect when assessing, for example, readability features, whose formulas are often designed specifically for the English language~\cite{Antunes2019_AdequacyReadability}. The process of evaluating these two dimensions was conducted by the authors of this study independently, and discrepancies between assessments were later discussed until an agreement was reached.

\begin{figure}[ht]
    \centering
    \includegraphics[width=0.75\textwidth]{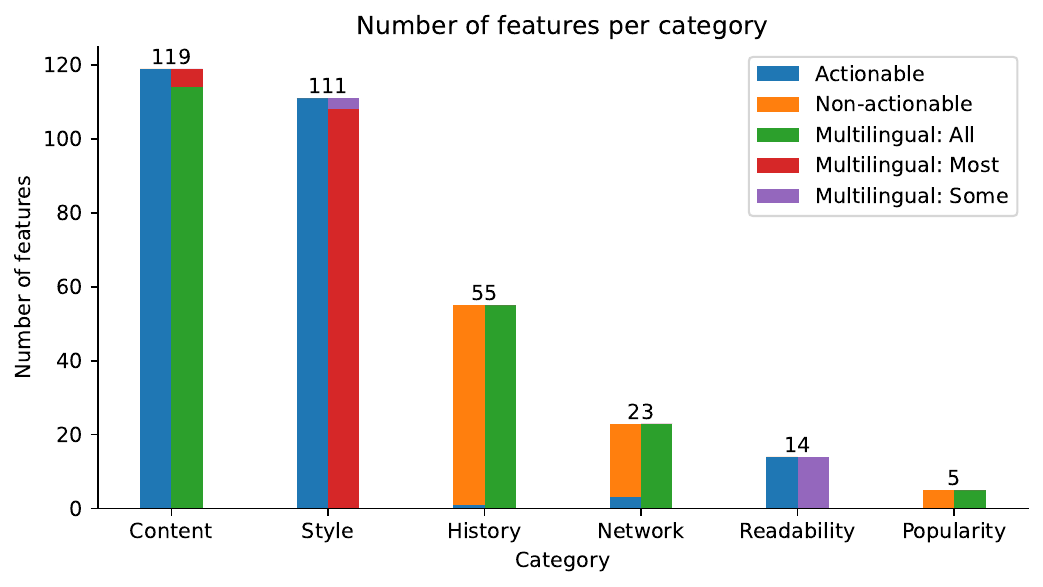}
    \caption{Collected Features: Count per Category}
    \label{fig:feature_count_graph}
\end{figure}

Overall, we collected \FEATURES\ distinct features throughout the \PUBTOTAL\ analyzed articles. Figure~\ref{fig:feature_count_graph} better displays the proportion of features per category and how they correlate to the other dimensions too. 

In this sub-section, we will overview every feature category, listing the 25\% most used features from each one (but never less than 15). We finish this sub-section by summarizing our findings.

\subsubsection{Content Features}

Content features contain information regarding the length and structure of the article. They are one of the simplest to compute since they can be almost directly derived from the text. Table~\ref{tab:feat_Content} lists the 25\% most used content features gathered from the literature review.

\input{sections/results/features/features.content}

Intuitively, there is a correlation between article length and quality. Good articles should not be too long and complex, overwhelming the reader, but not too simple either, as that could signify incomplete information. Also, according to Wikipedia~\cite{Web_WikipediaContentAss}, \emph{stub} articles (drafts) are ``usually very short'', which also demonstrates that correlation. 

Features related to the article structure are also essential to represent quality. Well-written articles should have a clear organization, with a balanced division of the content in sections and paragraphs. Images improve the reading experience, and references tend to increase the credibility of an article, so they are both decent indicators of quality.

\subsubsection{Style Features}

Style features evaluate how contributors write their sentences, measuring the used word classes and sentence types. Table~\ref{tab:feat_Style} provides the listing of the 25\% most used style features. The intuition behind these features is more subtle, but the idea is that certain language practices are considered of better quality when presenting information about a topic~\cite{Field2015_StyleGuide}.

\input{sections/results/features/features.style}

\subsubsection{Readability Features}

Readability features assess how easily readable a Wikipedia article is. They are calculated using carefully designed formulas that combine word, syllable, and character count and are listed in Table~\ref{tab:feat_Readability}. We describe the formula for each one below, on Equations~\ref{eq:ARI} to~\ref{eq:SG}. 

\vspace{1em}
\textbf{Automated Readability Index}: Estimates readability by combining the average word length with the average sentence size.
    
\begin{equation} 
    \label{eq:ARI}
    ARI = 4.71\frac{characters}{words} + 0.5\frac{words}{sentences} - 21.43
\end{equation}
\vspace{1em}

\textbf{Coleman-Liau}: Similarly to \(ARI\), estimates readability by combining the average word length with the average sentence size.
    
\begin{equation} 
    \label{eq:CL}
    CL = 5.88\frac{characters}{words} - 29.6\frac{sentences}{words} - 15.8
\end{equation}

\vspace{1em}
\textbf{Difficult Word Score (DWS)}: The DWS is calculated by counting the number of \emph{difficult words}, which is a definition that varies between papers. According to Dang \& Ignat, for example, ``A word is considered difficult if it does not appear in a list of 3000 common English words that groups of fourth-grade American students could reliably understand.''~\cite{Dang2016_lr16}.

\vspace{1em}
\textbf{Dale-Chall}: Also uses the concept of \emph{difficult words}, combining it with the average sentence size to estimate readability.
    
\begin{equation} 
    \label{eq:DC}
    DC = 0.1579 * (\frac{difficultwords}{words} * 100) + 0.0496\frac{words}{sentences}
\end{equation}

\vspace{1em}
\textbf{Flesch Reading Ease}: Using the average sentence size and amount of syllables per word, computes a value between 0 and 100, where 0 indicates the text is difficult to understand.
    
\begin{equation} 
    \label{eq:FRE}
    FRE = 206.835 - 1.015\frac{words}{sentences} - 84.6\frac{syllables}{words}
\end{equation}

\vspace{1em}
\textbf{Flesch-Kincaid Score}: Same as \(FRE\), but provides US grade levels instead of values between 0 and 100.
    
\begin{equation} 
    \label{eq:FK}
    FK = 0.39\frac{words}{sentences} - 11.8\frac{syllables}{words} - 15.59
\end{equation}

\vspace{1em}
\textbf{FORCAST Readability Formula}: Measures grade level from the number of monosyllabic words in a text sample of 150 words.

\begin{equation} 
    \label{eq:FOR}
    FOR = 20 - \frac{monosyllabic}{10}
\end{equation}

\vspace{1em}
\textbf{Gunning Fog Index}: Uses the concept of \(complexwords\), which is the number of words with three or more syllables. The higher its value, the more difficult is the text to read.
    
\begin{equation} 
    \label{eq:GFI}
    GFI = 0.4(\frac{words}{sentences} + 100\frac{complexwords}{words})
\end{equation}

\vspace{1em}
\textbf{Lasbarhets Index (LIX)}: Very similar to \(GFI\). In this case, \(complexwords\), is the number of words with more than six characters. The higher its value, the more difficult is the text to read.  
    
\begin{equation} 
    \label{eq:LBI}
    LIX = \frac{words}{sentences} + 100\frac{complexwords}{words}
\end{equation}

\vspace{1em}
\textbf{Linsear Write Formula}: Let \(n_1\) be the number of words with two syllables or less, and \(n_2\) be the number of words with three syllables or more.
    
\begin{equation} 
    \label{eq:LWF}
    LWF=\begin{cases}
        \frac{n_1 + 3 \times n_2}{sentences \times 2}, & \text{if \(\frac{n_1 + 3 \times n_2}{sentences} > 20 \)}.\\
        \frac{n_1 + 3 \times n_2}{sentences \times 2} - 1, & \text{otherwise}.
    \end{cases}
\end{equation}

\vspace{1em}
\textbf{Miyazaki Readability Score}: Outputs a result between 0 and 100. The higher its value, the more difficult is the text to read.

\begin{equation} 
    \label{eq:MYZ}
    MYZ = 164.935 - (18.792 * \frac{letters}{words} + 1.916\frac{words}{sentences})
\end{equation}

\vspace{1em}
\textbf{Smog-Grading}: \(polysyllables\) is the average number of polysyllabic words per 30 sentences (excluding proper names). They are usually calculated from a sample of 30 sentences.
    
\begin{equation} 
    \label{eq:SG}
    SG = 3 + \sqrt{polysyllables}
\end{equation}

\vspace{1em}
\textbf{Wiener Sachtextformel}: The authors propose multiple formulas, but always aim to measure the grade level required to understand a German text~\cite{Bamberger1984_WSachtextformel}.

\input{sections/results/features/features.readability}

\subsubsection{History Features}

History (or Review) features evaluate quality by analyzing an article's review history, authors, and contributions. The intuition behind them is that stable Wikipedia articles with trustworthy authors tend to be of better quality than controversial articles with very frequent edits. Table~\ref{tab:feat_History} lists the 25\% most mentioned history features, some of which make use of the concepts of \textbf{Active review}: Review made by one of the most active 5\% reviewers, and \textbf{Occasional review}: Review made by a user that edited the article less than four times.

\input{sections/results/features/features.history}

\subsubsection{Network features}

Network features are based on the articles' graph, in which nodes represent articles and the edges show the links between them. High-quality articles tend to cite other sources and are more likely to be cited by others. Therefore, these features may be strong indicators of quality, and we list the 15 most used ones in Table~\ref{tab:feat_Network}.

\input{sections/results/features/features.network}

\subsubsection{Popularity Features}

Popularity features are significantly less typical than the rest, as they require analytics information about the article. Still, ideally, high-quality articles would be more visited than worse ones. Table~\ref{tab:feat_Popularity} lists all the popularity features we collected.

\input{sections/results/features/features.popularity}

\subsubsection{Summary}

Most analyzed publications (\PUBFEATURES\ out of \PUBTOTAL) suggest or use article features for quality assessment. From those, we identified \ALLFEATURES\ features, \FEATURES\ unique. However, some papers do not always exhaustively detail the used features, so the real number may be much higher. For instance, Flekova et al.~\cite{Flekova2014_lr36} claim to use 3,000, but we only managed to identify 25. Still, some authors focus much more on feature discovery than others. Bassani and Viviani~\cite{Bassani2019_lr359}, and Anderka et al.~\cite{Anderka2012_lr17} both suggest more than 100 features in the ML solutions they propose, and ten papers suggest over 50~\cite{Dalip2014_lr1004, Wang2019_lr74, Magalhaes2019_lr2028, Dalip2011_lr1003, Wang2020_lr26, Dalip2009_lr14, Ferretti2012_lr115, Pereyra2019_lr147, Anderka2012_lr17, Dalip2014_lr1004}.

We can also note a correlation between a feature's category and its suitability for multilingual or actionable models. With a few exceptions, that pattern generally falls into what is listed in Table~\ref{tab:category_vs_actml}.

\begin{table}[htbp]
    \caption{Feature category vs Actionable and Multilingual properties}
    \label{tab:category_vs_actml}
    \centering
    \begin{tabular}{c c c}
        \toprule
        \textbf{Category} & \textbf{Actionable} & \textbf{Multilingual} \\
        \midrule
        Content & Yes & All \\
        Style & Yes & Most \\
        Readability & Yes & Some \\
        History & No & All \\
        Network & No & All \\
        Popularity & No & All \\
        \bottomrule
    \end{tabular}
\end{table}

\begin{figure}[ht]
    \centering
    \includegraphics[width=0.75\textwidth]{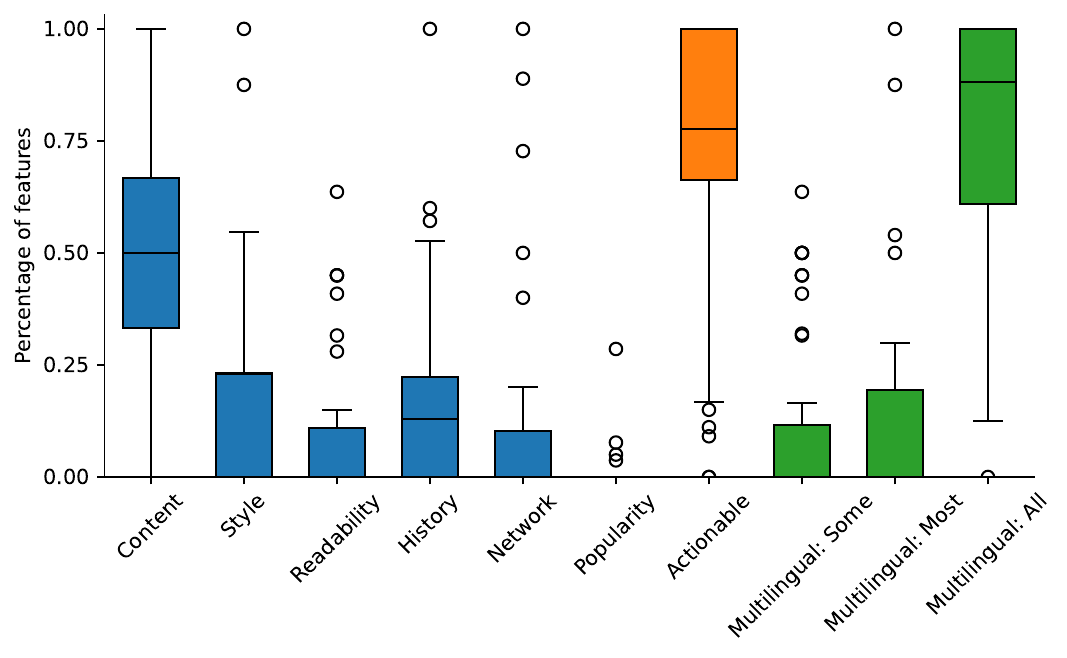}
    \caption{Feature characteristics within machine learning approaches (Blue - Feature Category, Orange - Actionable, Green - Multilingual). The y-axis shows the percentage of the respective features among all features used in the paper.}
    \label{fig:featuresml}
\end{figure}

Finally, we decided to assess the contexts in which the collected features are used to create machine learning models. Figure~\ref{fig:featuresml} overviews the typical feature sets in those solutions, showing which categories are more common, and whether papers tend to use actionable and multilingual features. Content features are clearly the most prevalent, but otherwise, there is no strong preference for a specific category. We do see a preference for multilingual and actionable features in these models, which indicates that the existing literature may be useful to authors who wish to further explore these topics.

\subsection{Quality Metrics}

Sometimes authors define new metrics, using them as direct measures of quality or as input for machine learning algorithms. We found that \PUBMETRICBASED\ of \PUBTOTAL\ included results use a direct metric-based approach, but \PUBMETRICS\ use metrics in their study, in some way.
This section describes the two most common types of metrics within the literature: Stvilia's IQ metrics, and Text Survival metrics.

\subsubsection{Stvilia's IQ Metrics}

Stvilia et al.~\cite{Stvilia2005_lr12, Stvilia2007_lr1012, Stvilia2005_lr1013} propose 7 Information Quality (IQ) metrics which, combining different article features, aim to evaluate Wikipedia quality more systematically. We took the definitions for each metric directly from their study~\cite{Stvilia2007_lr1012} and define their formulas in Equations~\ref{eq:stv_auth} to~\ref{eq:stv_vol}.

\textbf{Authority}: Authority is defined as ``the degree of the reputation of an information object in a given community''. \(Connectivity\) corresponds to the number of articles with at least one contributor in common with the assessed article.

\begin{multline} 
    \label{eq:stv_auth}
    Authority = 0.2 \times Unique Editors + 0.2 \times Contributions + 0.1 \times Connectivity + 0.3 \times Reverts + \\
    0.2 \times External Links + 0.1 \times Registered Contributions + 0.2 \times Anonymous Contributions
\end{multline}

\vspace{0.5cm}

\textbf{Completeness}: Authors define Completeness as ``the granularity or precision of an information object's model or content values''.

\begin{equation} 
    \label{eq:stv_comp}
    Completeness = 0,4 \times Internal Broken Links + 0,4 \times Internal Links + 0,2 \times Article Length
\end{equation}

\vspace{0.5cm}

\textbf{Complexity}: Complexity is defined as ``the degree of cognitive complexity of an information object relative to a particular activity''.

\begin{equation} 
    \label{eq:stv_compx}
    Complexity = 0,5 \times Flesch reading Ease - 0,5 \times Flesch Kincaid grade level
\end{equation}

\vspace{0.5cm}

\textbf{Informativeness}: Measures the amount of information in a document. \(InfoNoise\) represents the ratio between the size of the information and the article, measuring the amount of \emph{noise} in the document. \(Diversity\) refers to the ratio between editors and total edits.

\begin{equation} 
    \label{eq:stv_info}
    Informativeness = 0,6 \times InfoNoise - 0,6 \times Diversity + 0,3 \times Images
\end{equation}

\vspace{0.5cm}

\textbf{Consistency}: Consistency is defined as ``the extent to which similar attributes or elements of an information object are consistently represented with the same structure, format, and precision''.

\begin{equation} 
    \label{eq:stv_cons}
    Consistency = 0,6 \times Administrators Edit Share + 0,5 \times Age_{days}
\end{equation}

\vspace{0.5cm}

\textbf{Currency}: Currency corresponds to ``the age of an information object'' in days.

\begin{equation} 
    \label{eq:stv_curr}
    Currency = Collection Date - Last Edit Date
\end{equation}

\vspace{0.5cm}

\textbf{Volatility}: Volatility measures ``the amount of time the information remains valid''.
\begin{equation} 
    \label{eq:stv_vol}
    Volatility = Median Revert Time
\end{equation}

\vspace{0.5cm}

Multiple authors talk about and experiment with these metrics \cite{Warncke-Wang2013_lr13, Sugandhika2022_lr119, Couto2021_lr145, Couto2021_lr161, Lee2013_lr202}, although rarely within a Machine Learning context. The application of Stvilia's metrics to training ML algorithms could be worthy of experimentation.

\subsubsection{Text Survival}

Most metric-based approaches rely on the idea of \textit{text survival}: If a piece of the article survives many revisions, that part is likely of good quality. The most common examples are \textit{ProbReview} and \textit{PeerReview}, initially proposed by Hu et al.~\cite{Hu2007_lr1, Hu2007_lr58}. Many authors use \textit{ProbReview}~\cite{Dalip2009_lr14, Wang2020_lr26, Wang2019_lr74, Bassani2019_lr359, Dalip2016_lr1002, Dalip2011_lr1003, Dalip2014_lr1004, Dalip2012_lr2014, Magalhaes2019_lr2028} and \textit{PeerReview}~\cite{Robertie2015_lr32, Wang2019_lr74, Suzuki2012_lr117} in their works, but \textit{text survival} is a frequent concept in metric-based approaches. For example, the proposed measures of \textit{Trensient Contribution} and \textit{Persistent Contribution}~\cite{Wohner2009_lr10, Wang2010_lr70}, \textit{Word Persistence}~\cite{Halfaker2009_lr19}, among others~\cite{Suzuki2012_lr60, Suzuki2013_lr79, Qin2012_lr131, Suzuki2013_lr133, Suzuki2012_lr159, Robertie2017_lr172} also rely on that notion.

%% file: sections/results/features/features.content.tex
\begin{table}[htbp]
    \caption{List of 25\% most used Content features mentioned in the assessed publications (complete feature set and paper citations in our dataset, file 'Full Feature List.pdf')}
    \label{tab:feat_Content}
    \centering
    \begin{tabular}{m{0.5\textwidth} c c c}
        \toprule
        \textbf{Name} & \textbf{Actionable} & \textbf{Multilingual} & \textbf{\# Papers} \\ 
        \midrule
        Character Count (Article Length) & Yes & All & 61 \\
        Internal Link Count & Yes & All & 44 \\
        Image Count & Yes & All & 42 \\
        External Link Count & Yes & All & 35 \\
        Word Count & Yes & All & 32 \\
        Section Count & Yes & All & 32 \\
        Reference Count & Yes & All & 32 \\
        Subsection Count & Yes & All & 24 \\
        Sentence Count & Yes & All & 22 \\
        Category Count & Yes & All & 21 \\
        Citation Count & Yes & All & 17 \\
        Information Noise Score (InfoNoise) (multiple definitions) & Yes & All & 16 \\
        Longest Sentence Length (Words) & Yes & All & 14 \\
        Has InfoBox & Yes & All & 14 \\
        Subsection Count per Section (section nesting) & Yes & All & 13 \\
        Mean Sentence Length (Words) & Yes & All & 12 \\
        Paragraph Count & Yes & All & 12 \\
        Image Count per Character & Yes & All & 12 \\
        Image Count per Section & Yes & All & 12 \\
        Heading Count & Yes & All & 12 \\
        Mean Paragraph Length in Words & Yes & All & 11 \\
        Reference Count per Character & Yes & All & 11 \\
        Introduction Length (Lead section; abstract) & Yes & All & 10 \\
        Mean Section Length in Words & Yes & All & 9 \\
        Citation Count per Section & Yes & All & 9 \\
        Table Count & Yes & All & 9 \\
        Mean Word Length & Yes & All & 8 \\
        Section Length Stdev. & Yes & All & 8 \\
        Citation Count per Character & Yes & All & 8 \\
        \bottomrule
    \end{tabular}
\end{table}

%% file: sections/results/features/features.style.tex
\begin{table}[htbp]
    \caption{List of 25\% most used Style features mentioned in the assessed publications (complete feature set and paper citations in our dataset, file 'Full Feature List.pdf')}
    \label{tab:feat_Style}
    \centering
    \begin{tabular}{m{0.5\textwidth} c c c}
        \toprule
        \textbf{Name} & \textbf{Actionable} & \textbf{Multilingual} & \textbf{\# Papers} \\ 
        \midrule
        Short Sentence Rate (multiple definitions)  & Yes & Most & 14 \\
        Long Sentence Rate (multiple definitions) & Yes & Most & 13 \\
        Passive Voice Sentence Count & Yes & Most & 10 \\
        Question Count & Yes & Most & 9 \\
        Auxiliary verb count & Yes & Most & 9 \\
        Number of sentences starting with a pronoun & Yes & Most & 8 \\
        To be Verb Count per Word & Yes & Most & 8 \\
        Coordinate Conjunction Count per Word & Yes & Most & 8 \\
        Number of sentences starting with an article & Yes & Most & 7 \\
        Number of sentences starting with a coordinate conjunction & Yes & Most & 7 \\
        Number of sentences starting with a subordinate preposition or conjunction & Yes & Most & 7 \\
        Pronoun Count & Yes & Most & 7 \\
        Preposition Count per Word & Yes & Most & 7 \\
        Character/PoS N-grams & Yes & Some & 7 \\
        Syllable Count & Yes & Most & 6 \\
        Number of sentences starting with an interrogative pronoun & Yes & Most & 6 \\
        Nominalization Count per Word & Yes & Most & 6 \\
        Number of sentences starting with a preposition & Yes & Most & 5 \\
        To be Verb Count & Yes & Most & 5 \\
        Long Words Rate (multiple definitions) & Yes & Most & 4 \\
        Question Count per Sentence & Yes & Most & 4 \\
        Passive Voice Sentence Count per Sentence & Yes & Most & 4 \\
        Syllable Count per Word & Yes & Most & 4 \\
        One-Syllable Word Count & Yes & Most & 4 \\
        Number of sentences starting with a pronoun per Sentence & Yes & Most & 4 \\
        Number of sentences starting with an article per Sentence & Yes & Most & 4 \\
        \bottomrule
    \end{tabular}
\end{table}

%% file: sections/results/features/features.readability.tex
\begin{table}[htbp]
    \caption{List of all Readability features mentioned in the assessed publications (complete feature set and paper citations in our dataset, file 'Full Feature List.pdf')}
    \label{tab:feat_Readability}
    \centering
    \begin{tabular}{m{0.5\textwidth} c c c}
        \toprule
        \textbf{Name} & \textbf{Actionable} & \textbf{Multilingual} & \textbf{\# Papers} \\ 
        \midrule
        Flesch-Kincaid & Yes & Some & 27 \\
        Flesch reading ease & Yes & Some & 24 \\
        Automated Readability Index & Yes & Some & 23 \\
        Coleman-Lieau & Yes & Some & 20 \\
        Smog-Grading & Yes & Some & 20 \\
        Gunning Fog Index & Yes & Some & 19 \\
        Lasbarhets Index (LIX) & Yes & Some & 13 \\
        Dale-Chall & Yes & Some & 10 \\
        Linsear Write Formula & Yes & Some & 6 \\
        Difficult Word Score & Yes & Some & 5 \\
        Bormuth Readability Index & Yes & Some & 4 \\
        Miyazaki Readability Score & Yes & Some & 4 \\
        FORCAST Readability Formula & Yes & Some & 3 \\
        Wiener Sachtextformel & Yes & Some & 1 \\
        \bottomrule
    \end{tabular}
\end{table}

%% file: sections/results/features/features.history.tex
\begin{table}[htbp]
    \caption{List of 15 most used History features mentioned in the assessed publications (complete feature set and paper citations in our dataset, file 'Full Feature List.pdf')}
    \label{tab:feat_History}
    \centering
    \begin{tabular}{m{0.5\textwidth} c c c}
        \toprule
        \textbf{Name} & \textbf{Actionable} & \textbf{Multilingual} & \textbf{\# Papers} \\ 
        \midrule
        Revision Count & No & All & 39 \\
        Contributor Count & No & All & 39 \\
        Article Age (days) & No & All & 34 \\
        Discussion Count & No & All & 18 \\
        Anonymous Contributor Count & No & All & 15 \\
        Revisions per Contributor & No & All & 13 \\
        Revisions per Day & No & All & 12 \\
        Registered Contributor Count & No & All & 12 \\
        Registered Revision Count & No & All & 12 \\
        Anonymous Revision Count & No & All & 10 \\
        Current Revision Age (days) (currency) & No & All & 9 \\
        Revisions per Contributor Stdev. & No & All & 9 \\
        Recent Revision Count per Revision & No & All & 9 \\
        Occasional Revision Count per Revision  & No & All & 8 \\
        Article Age per Revision & No & All & 7 \\
        \bottomrule
    \end{tabular}
\end{table}

%% file: sections/results/features/features.network.tex
\begin{table}[htbp]
    \caption{List of 15 most used Network features mentioned in the assessed publications (complete feature set and paper citations in our dataset, file 'Full Feature List.pdf')}
    \label{tab:feat_Network}
    \centering
    \begin{tabular}{m{0.5\textwidth} c c c}
        \toprule
        \textbf{Name} & \textbf{Actionable} & \textbf{Multilingual} & \textbf{\# Papers} \\ 
        \midrule
        PageRank & No & All & 20 \\
        Incoming Internal Link Count & No & All & 16 \\
        In-degree & No & All & 15 \\
        Out-degree & Yes & All & 14 \\
        Local Clustering Coefficient & No & All & 14 \\
        \# of Versions in other Languages & Yes & All & 14 \\
        Reciprocity within Neighbors & No & All & 12 \\
        Assortativity (in-in, in-out, out-in, out-out) & No & All & 9 \\
        Number of Articles with Common Editors (Connectivity) & No & All & 4 \\
        Betweenness Centrality & No & All & 3 \\
        Neighbor Count & Yes & All & 2 \\
        Average Path Length & No & All & 2 \\
        K-Core Number & No & All & 2 \\
        HITS (Hyperlink-Induced Topic Selection) & No & All & 2 \\
        Density & No & All & 1 \\
        \bottomrule
    \end{tabular}
\end{table}

%% file: sections/results/features/features.popularity.tex
\begin{table}[htbp]
    \caption{List of all Popularity features mentioned in the assessed publications (complete feature set and paper citations in our dataset, file 'Full Feature List.pdf')}
    \label{tab:feat_Popularity}
    \centering
    \begin{tabular}{m{0.5\textwidth} c c c}
        \toprule
        \textbf{Name} & \textbf{Actionable} & \textbf{Multilingual} & \textbf{\# Papers} \\ 
        \midrule
        Number of Page Visits & No & All & 3 \\
        Number of Page Watchers & No & All & 3 \\
        Number of Page Visits per Day & No & All & 1 \\
        Number of Page Viewers & No & All & 1 \\
        Number of Shares in Social Media & No & All & 1 \\
        \bottomrule
    \end{tabular}
\end{table}

%% file: sections/section7-discussion.tex
\section{Discussion} \label{sec:discussion}

This section discusses the results we obtained while attempting to answer the research questions we stated in Section~\ref{sec:method}.

\subsection{RQ1. \RQONE}

We can identify three typical quality prediction strategies: metric-based approaches, classical machine learning models trained with article features, and deep learning methods using full text or features. Papers with other focus occasionally show up too. For example, Shen et al.~\cite{Shen2019_lr1061, Shen2020_lr2009} proposed a multimodal classifier that uses both article features and a visual rendering of the document as input for quality prediction. There are also studies about quality flaw analysis and prediction~\cite{Anderka2012_lr17, Anderka2011_lr35, Ferschke2012_lr43, Wang2021_lr48, Ferretti2018_lr100, Pereyra2019_lr147, Urquiza2016_lr160, Anderka2011_lr1027, Li2022_lr2019}, which identify frequent patterns of improvement.

We would also like to highlight Halfaker and Geiger's study~\cite{Halfaker2020_lr1055}, where they propose ORES: an API-based service that supports real-time scoring of Wikipedia edits, supporting many languages, and achieving exceptional results. The study has had a great impact in this research area and the service is currently provided by Wikimedia~\footnote{ORES API: ttps://ores.wikimedia.org/}, setting a great benchmark for all future work.

It is challenging to determine the best strategies, as each one has its advantages and drawbacks. Metric-based approaches do not require model training, and some deep learning solutions are difficult to apply in a real-time scenario (e.g., Dang \& Ignat's~\cite{Dang2017_lr23}). Regardless, our study showed that every strategy can perform effectively with the proper configuration (6-class: >60\% accuracy, 2-class: >95\% accuracy), and it is up to the researchers/developers to determine which solution makes the most sense for their context.

\subsection{RQ2. \RQTWO}

Authors experiment with many machine learning algorithms, totaling \MLEXPERIMENTS\ distinct experiments. It is not trivial to decide which algorithm is the most effective, as that is essentially dependent on the dataset definitions, but we see great performances from solutions using LSTMs, GRUs, Random Forests, and SVMs. Furthermore, boosting strategies, which combine multiple weak models into stronger ones~\cite{Schapire2003_Boosting} also tend to be very effective with classical algorithms (e.g., Decision Trees).

We have shown that most studies formulate this problem as a classification task, but we would like to note Teblunthuis's work~\cite{Teblunthuis2021_ExtendingORES}, which shows that the English Wikipedia's quality levels (FA's, GA's, Stubs, etc.) are not evenly distributed on a linear scale, proposing a spacing that more effectively represents how distant are the multiple levels of quality. We did not include the study in this review, because it does not directly assess the quality of Wikipedia articles, but is still an insightful analysis of Wikipedia's quality scale.

Deep Learning is a topic that has gained significant popularity during this decade~\cite{Yapici2019_DLReview}. Most solutions we reviewed use classical approaches (e.g., decision trees, SVM), but we also discovered multiple deep learning solutions (e.g., LSTM). Additionally, we noticed that deep learning approaches were almost nonexistent ten years ago, while studies using classical methods for the automatic assessment of Wikipedia are not as common these days, relatively speaking.

Overall, although classical statistical learning approaches (e.g., decision trees, SVM) are more common, there has been a notable recent preference for deep learning (e.g., LSTM). Additionally, we noticed that deep learning approaches were almost nonexistent ten years ago, while studies using classical methods for the automatic assessment of Wikipedia are not as common nowadays, in relative terms. It is dangerous to directly compare results but, so far, deep learning seems to show slightly better performance than previous solutions. Furthermore, deep learning has been gaining significant popularity during the past decade~\cite{Yapici2019_DLReview} and, if this trend continues, it is possible that their performance eclipses the effectiveness of classical algorithms.

\subsection{RQ3. \RQTHREE}
 
In this review, we collected \FEATURES\ different features, each of them factoring the text of an article, its review history, how it relates to other articles within Wikipedia, or even its popularity within users.

Even though Style features are the second largest group of identified features, they are one of the least used categories. Content features, which consider only the length and structure of the article, are both the most abundant and the most frequently used ones. The other features also appear often but not as much, possibly due to their higher computation complexity.

Not all papers use simple article features to assess quality, though. Some deep learning models train with the articles' full texts, and other studies opt for a metric-based approach, as shown in Section~\ref{sec:res-ml}, but these approaches are not as common. Besides, although they may be used with Machine Learning, metrics are better suited for more manual approaches, so it is not surprising they do not show up as often in this review.

Some papers also compare different subsets of features regarding their effectiveness at predicting quality~\cite{Dalip2009_lr14, Dang2016_lr16, Wang2020_lr26, Flekova2014_lr36, Ferschke2012_lr43, Wang2019_lr74, Dalip2011_lr1003, Dalip2014_lr1004}. By analyzing the different studies, we see that Content and Style features appear to be the most effective, but History and Network features are sometimes considered very useful for predicting quality too, so it would be wise to combine all categories. Readability Features also appear to generate decent results, but not so significantly as the others do, since those already combine existing features in a predefined way. 



\subsection{RQ4. \RQFOUR}

\begin{figure}[htbp]
    \centering
    \includegraphics[width=0.7\textwidth]{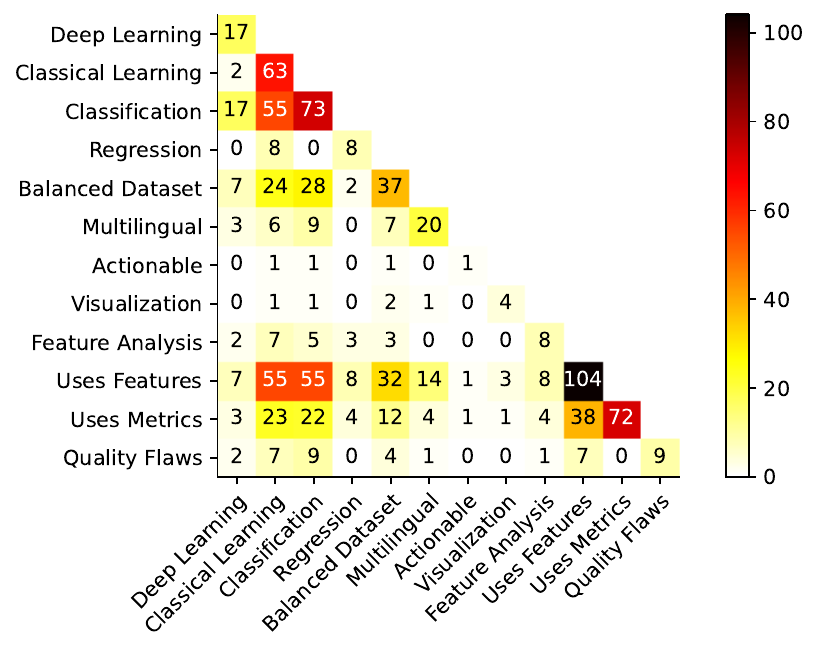}
    \caption{Number of studies incorporating different methodological aspects}
    \label{fig:gapmatrix}
\end{figure}

As the basis for a gap analysis, we organized multiple recurring methodology aspects into a frequency matrix, represented by the heatmap in Figure~\ref{fig:gapmatrix}. Since we only pair the items two by two, this does not give us a picture of all the possible methods, but still provides quite some insight into what authors explore more and less within this field.

We can initially notice that the research within machine learning and feature-based models is extensive. There is also some work on metric-based solutions and studies focused on multiple Wikipedia languages. Nonetheless, there still exist some areas which the literature does not cover as much.

The most notable gap concerns the actionable solutions we previously discussed. Ideally, a model would not only predict article quality but also suggest possible steps for improvement. Warncke-Wang~\cite{Warncke-Wang2013_lr13} proposes such a model, but the focus on actionable models seems to be otherwise scarce within the existing literature. Some studies~\cite{Wecel2015_lr34, Chevalier2010_lr65, Sciascio2017_lr83, Dalip2011_lr111} propose reporting and visualization tools for analyzing the quality of Wikipedia, thus somewhat exploring this concept, but there is a clear lack of studies concerned about the topic. This review distinguishes actionable and non-actionable features in Section~\ref{sec:res-features}, aiming to guide authors in studying tools that assist Wikipedia readers and editors.

There is also very little work on multilingual solutions using machine learning, and none of those experiments with regression models. In addition, in studies that do design multilingual solutions, non-English model performance rarely comes close to the English one. For instance, the ORES study~\cite{Halfaker2020_lr1055} predicts the quality of English articles with an accuracy of 62.9\% but the French model's performance is only 44.2\%. Wikipedia has millions of visitors using hundreds of versions~\cite{Web_WikipediaList}, and as we discussed in Section~\ref{sec:res-features}, not every concept makes sense in every language, so it is crucial to study approaches that perform well in multiple languages.

There are some other issues not represented in the heatmap. For instance, there could also be a need for the existence of quality models specialized for specific Wikipedia categories. A couple of studies conduct several experiments with a few categories (e.g., history, biology, health), and some actually make models directed for specific fields~\cite{Flekova2014_lr36, Couto2021_lr161}, but there is little work on models specializing on more than one area simultaneously.


Finally, we were surprised not to see a lot of concern for the reproducibility and replicability~\cite{Web_Reproducibility} of the studies. Only ten papers share the source code of their work (from what we discovered, at least), and shared datasets are often inaccessible. We also sometimes struggled to locate even a description of the class distribution of the datasets, which is essential when comparing machine learning results. Hopefully, future authors will give increased importance to making their work more reproducible.

In summary, the quality assessment of Wikipedia is a significantly researched topic, for which there are many diverse methods to approach it. The results we collected will help any future work related to this field, and further experimentation may help develop better quality predictors.

%% file: sections/section8-conclusion.tex
\section{Conclusions and Future Work} \label{sec:conclusions}

This study reviewed literature related to the automatic assessment of the quality of Wikipedia articles, performing an in-depth analysis of 149 different papers out of thousands of inspected results. Our findings indicate that research on this topic has fluctuated for the past few years but only started getting attention several years after the launch of Wikipedia. There are many different proposals, but most use a feature-based traditional machine learning approach and refer to Wikipedia's content assessment standards to measure quality. We are starting to see more focus on deep learning methods, which may soon become the definite best option for this task, but it is difficult to compare results directly, since performance metrics, number of classes, and label distribution vary from study to study.

We can identify some limitations in our study, though. The most notable is the lack of non-bibliographic sources within our selection. Although our methodology should cover most journal submissions, conference papers, and other research repositories, some relevant studies may still be missed, such as Johnson's proposed quality model~\cite{Web_WikimediaModel}. Nevertheless, nearly all relevant publications should be accessible through standard digital libraries. 

Upon reviewing so much literature about the topic, we were puzzled by the fact that automatic assessment methods are still not widely used in Wikipedia. Although it is difficult to produce a direct answer, there are multiple potential explanations:

\begin{enumerate}[itemsep=0pt]
    \item Reliability: Even though some machine learning methods show impressive performance, model accuracies are far from 100\%. This should not be a major issue, though, considering that, apart from B/C-tier articles, some papers show almost perfect one-off accuracy results.
    \item Complexity: As we have discussed before, quality is an extremely intricate concept with numerous properties, and some of them are much more challenging to assess than others. For instance, distinguishing a well-structured article from a poorly-structured one is trivial, compared to detecting false statements in a paragraph. Although this study does not focus so much on the trustworthiness part of information quality, all quality properties are relevant to Wikipedia users. As such, tools that do not fully grasp the essence of information quality may not be so well-received by the community.
    \item Accessibility: As we discussed in Section~\ref{sec:discussion}, there are not many reporting and visualization tools available for multilingual purposes, and we have seen that models are not easily transferable to other languages. We do have ORES~\footnote{ORES API: \url{https://ores.wikimedia.org/}}~\cite{Halfaker2020_lr1055} and WikiRank~\footnote{WikiRank website: \url{https://wikirank.net/}}~\cite{Wecel2015_lr34}, but ORES is an API service directed to editors and WikiRank only provides a few actionable items for improvement. Besides, without a more direct integration with Wikipedia, it is difficult for a casual Wikipedia user to learn about those tools and know how to handle them.
    \item Self-regulation: Unlike most social media, Wikipedia has no central authority, and instead relies on collaborative moderation so, by design, it cannot have a ground-truth. This is why Wikimedia is reluctant to apply AI moderation to the website~\cite{Web_WikimediaDisinformation}, and may also explain why automatic quality assessment methods are not as prevalent within Wikipedia.
\end{enumerate}

We cannot solve all these impediments but we believe there is potential in combining existing approaches and making a tool accessible to every Wikipedia user, providing instantaneous feedback concerning the quality of the article. Such a project could promote the widespread usage of automatic Wikipedia quality models, and the results of this review are helpful indicators of which techniques lead to better performance. Still, future researchers must design and conduct their own set of experiments, comparing languages, features, metrics, and datasets, as model performance depends on much more than just its algorithm.

At the time of writing, OpenAI's GPT-4 has just been released~\cite{OpenAI2023_GPT4}, and the major tech companies are racing to compete against the novel ChatGPT~\cite{Web_AIRace}. Right now, the previously inconceivable idea of using a chatbot to predict Wikipedia article quality, explain its reasoning, and suggest items for improvement, sounds much more feasible. This would require further research and development, and it is impossible to predict how this technology will advance in the near future, but it is evident how these tools could evolve to assist the topic of this research.